\documentclass[11pt, a4paper, logo, copyright]{googledeepmind}

\usepackage[authoryear, sort&compress, round]{natbib}
\usepackage[inkscapeformat=png]{svg}

\usepackage[most, breakable, skins]{tcolorbox}

\tcbuselibrary{skins}
\usepackage{lipsum}
\usepackage{tabularx}
\usepackage{afterpage}
\usepackage{booktabs}
\usepackage{subcaption}
\usepackage[frozencache,cachedir=minted-cache]{minted}
\usepackage{makecell}
\usepackage{multirow}
\usepackage{multicol}
\usepackage{array}
\usepackage{epigraph}
\usepackage{float}
\usepackage{listings, listings-rust}
\usepackage{fontawesome5}
\usepackage{amssymb,graphicx}
\usepackage[dvipsnames]{xcolor}
\usepackage{hyperref}
\usepackage[capitalise, noabbrev]{cleveref}
\usepackage{longtable}
\usepackage{graphicx}
\usepackage{pdflscape}
\usepackage{adjustbox}
\usepackage[section]{placeins}  % float barrier for each section
\usepackage{tikz}
\usetikzlibrary{shapes.geometric, arrows, positioning, fit}
\usepackage{tcolorbox} % For the shaded box around the prompt
\newcommand*{\colorboxed}{}
\def\colorboxed#1#{%
  \colorboxedAux{#1}%
}
\newcommand*{\colorboxedAux}[3]{%
  \begingroup
    \colorlet{cb@saved}{.}%
    \color#1{#2}%
    \boxed{%
      \color{cb@saved}%
      #3%
    }%
  \endgroup
}

\title{Michelangelo: Long Context Evaluations Beyond Haystacks via Latent Structure Queries}

\correspondingauthor{kirannv@google.com}


\author[*,1]{Kiran Vodrahalli}
\author[1]{Santiago Onta\~{n}\'{o}n}
\author[1]{Nilesh Tripuraneni}
\author[1]{Kelvin Xu}
\author[1]{Sanil Jain}
\author[1]{Rakesh Shivanna}
\author[1]{Jeffrey Hui}
\author[2]{Nishanth Dikkala}
\author[1]{Mehran Kazemi}
\author[2]{Bahare Fatemi}
\author[1]{Rohan Anil}
\author[1]{Ethan Dyer}
\author[1]{Siamak Shakeri}
\author[1]{Roopali Vij}
\author[1]{Harsh Mehta}
\author[1]{Vinay Ramasesh}
\author[1]{Quoc Le}
\author[1]{Ed Chi}
\author[1]{Yifeng Lu}
\author[1]{Orhan Firat}
\author[1]{Angeliki Lazaridou}
\author[1]{Jean-Baptiste Lespiau}
\author[1]{Nithya Attaluri}
\author[1]{Kate Olszewska}

\newcommand{\evalname}{Michelangelo\space}
\newcommand{\frameworkname}{Latent Structure Queries }
\newcommand{\frameworkshort}{LSQ}

\affil[*]{Corresponding author}
\affil[1]{Google DeepMind}
\affil[2]{Google Research}

\begin{abstract}
We introduce Michelangelo: a minimal, synthetic, and unleaked long-context reasoning evaluation for large language models which is also easy to automatically score. This evaluation is derived via a novel, unifying framework for evaluations over arbitrarily long contexts which measure the model's ability to do more than retrieve a single piece of information from its context. The central idea of the \frameworkname framework (\frameworkshort) is to construct tasks which require a model to ``chisel away'' the irrelevant information in the context, revealing a latent structure in the context. To verify a model's understanding of this latent structure, we query the model for details of the structure. Using \frameworkshort, we produce three diagnostic long-context evaluations across code and natural-language domains intended to provide a stronger signal of long-context language model capabilities. We perform evaluations on several state-of-the-art models and demonstrate both that a) the proposed evaluations are high-signal and b) that there is significant room for improvement in synthesizing long-context information.
\end{abstract}

\begin{document}

\maketitle

\epigraph{\textit{``The sculpture is already complete within the marble block, before I start my work. It is already there, I just have to chisel away the superfluous material.''}}{--- Michelangelo}

\section{Introduction}
\label{sec:intro}

The famous Renaissance sculptor Michelangelo was once asked how he was able to build such beautiful sculptures -- his answer, long immortalized as a quintessential comment on the nature of understanding and artistry, was as follows: ``The sculpture is already complete within the marble block, before I start my work. It is already there, I just have to chisel away the superfluous material.'' In this work, we draw a direct analogy between long-context language model understanding and the essence of the sculptor's task -- an extremely large context can be viewed as a block of marble with many irrelevant pieces of information, which the model must chisel away to reveal an understanding of the structure within. We introduce Michelangelo, an automatic diagnostic long-context understanding evaluation which is directly based on this principle, and develop the \frameworkname (\frameworkshort) framework for generating such evaluations. \frameworkshort\space allows us to develop considerably more complex evaluations as compared to the popular needle in the haystack methodology \citep{needle}. By requiring the model to extract information from structures rather than values from keys (sculptures from marble rather than needles from haystacks), we can more deeply test language model context understanding beyond retrieval.

\begin{figure}
    \centering
    \includegraphics[width=0.8\textwidth]{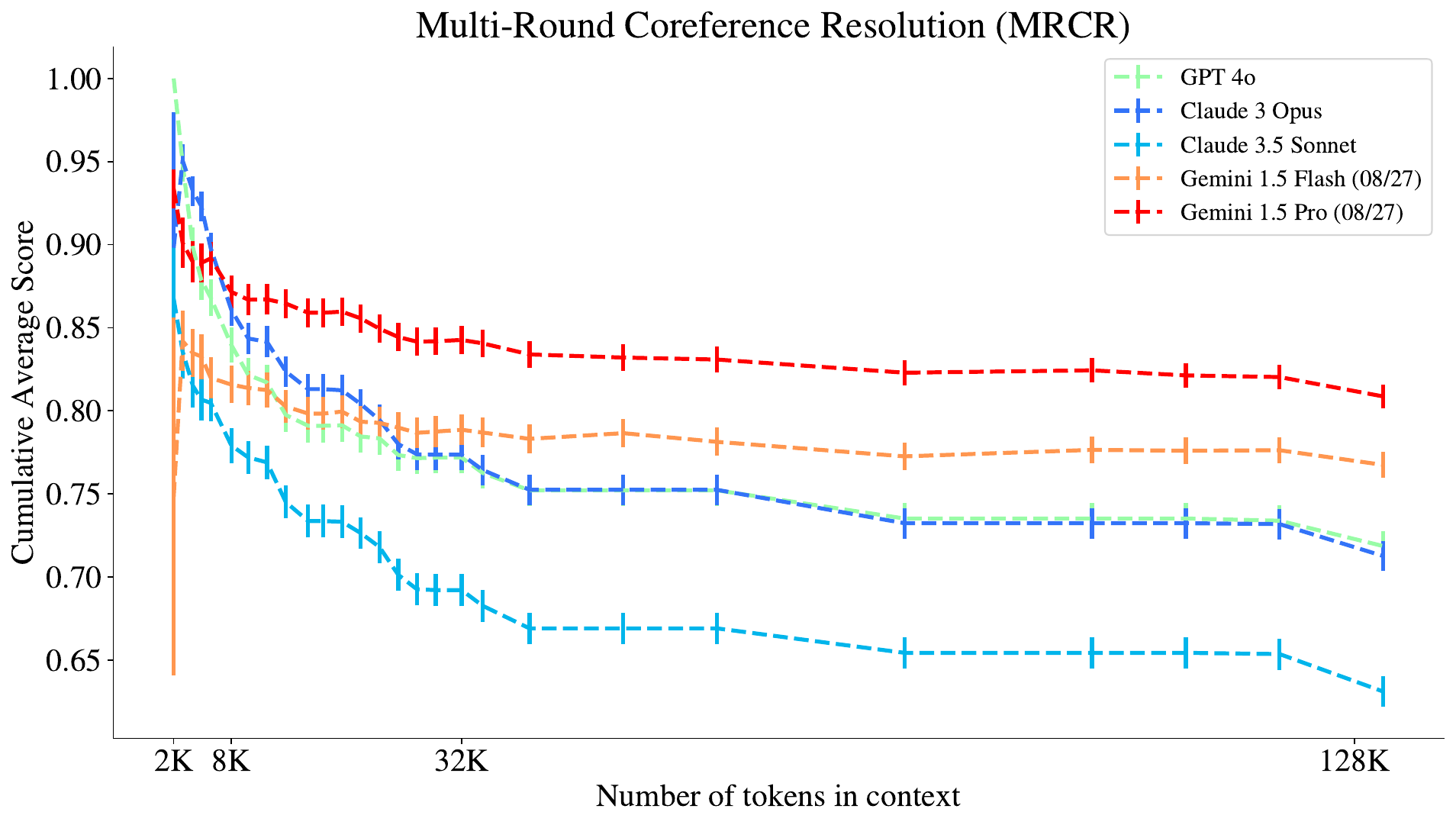}
    \caption{Performance of frontier models on the MRCR (multi-round coreference resolution) task, a synthetic long reasoning task. All models experience significant fall off in performance before 32K.}
    \label{fig:fig1}
\end{figure}

Large language models with extremely long context lengths from 128K tokens to surpassing 1M tokens \citep{claude, openai2023gpt4, geminiteam2024gemini15unlockingmultimodal} have recently become quite prominent in the literature. However, the problem of developing reliable tests to measure the extent to which these models are able to utilise the information in their context remains open.

The majority of attention in long-context evaluation has been focused on retrieval tasks, as popularized by the single-needle-in-a-haystack retrieval task \citep{needle}, with several follow-up works additionally examining the retrieval of multiple needles \citep{geminiteam2024gemini15unlockingmultimodal, li2024needlebenchllmsretrievalreasoning, hsieh2024rulerwhatsrealcontext, zhang2024infty}. Several more realistic long-context question-answering evaluations have also been developed \citep{kovcisky2018narrativeqa, bohnet2024longspanquestionansweringautomaticquestion, zhang2024infty}, and these tend to essentially reduce to solving a retrieval task in a more realistic setting.

More recently, several authors have highlighted the need for tests of long-context capability which go beyond retrieval \citep{goldman2024reallylongcontextneed, levy2024same, karpinska2024thousandpairsnovelchallenge}. These results point out that the ability of a model to retrieve one or more facts does not necessarily imply that a model is able to synthesize information from the full context. Existing benchmarks which attempt to measure a model's capacity for reasoning over a long context \citep{arora2023zoologymeasuringimprovingrecall, li2024needlebenchllmsretrievalreasoning} may have a few suboptimal properties. Some examples include:
\begin{itemize}
    \item relatively small context lengths;
    \item high artificiality with no natural language or code setting;
    \item the requirement of significant amounts of human labor to extend to longer context lengths;
    \item a construction that clearly avoids scenarios where the tasks require the usage of the full context, either because the information required to answer questions may be present in pretraining data or because it is possible to ``short-circuit'' the context length and answer the question with more local information.
\end{itemize}
Other approaches have focused on measuring a model's capability for performing many-shot learning \citep{agarwal2024manyshotincontextlearning, bohnet2024exploringbenchmarkingplanningcapabilities} and summarization \citep{chang2024booookscore, kim2024fablesevaluatingfaithfulnesscontent}, neither of which necessarily require or measure a model's capacity to reason over the context. Additionally, many existing long-context evaluation benchmarks construct their evaluation by utilizing pre-existing (and likely leaked-in-training-data) evaluations, making it difficult to ascertain the extent to which performance on these benchmarks may be due to leaked information \citep{bai2023longbench, li2024needlebenchllmsretrievalreasoning, hsieh2024rulerwhatsrealcontext, agarwal2024manyshotincontextlearning, bohnet2024exploringbenchmarkingplanningcapabilities, zhang2024infty, loft}. In this work, we focus on developing Michelangelo, a minimal, synthetic, and un-leaked long-context reasoning evaluation which is also extremely easy to automatically score.

\textbf{Minimality:}
\evalname constitutes a minimal set of the simplest canonical tasks which require understanding of the context beyond retrieval, and yet are still challenging for present-day models to solve. Minimal benchmarks, in both size and complexity, which capture a wide range of behavior are desirable for many reasons: they are easy to interpret, easy to use for tracking progress, and inform the simplest cases where a model is likely to fail.

\textbf{Canonical Primitives:}
\evalname constitutes three intuitive and simple long-context synthesis task primitives which require the model to synthesize multiple pieces of information strewn throughout the context in order to produce an answer, and which measure different aspects of a model's synthesis capabilities to provide a more holistic understanding of long-context model behavior. Each of the evaluations we present is positioned in a natural language or code-based setting, and is less synthetic compared to existing benchmarks. We observe that it is already challenging to synthesize two pieces of information across a context. In particular, \evalname measures the ability of a model to 
\begin{itemize}
    \item reproduce a given unique portion of the context while understanding ordering;
    \item understand a sequence of modifications to a list (which captures a very wide range of preexisting tasks);
    \item determine whether the answer to a query is contained within the context.
\end{itemize}

\textbf{Arbitrary Context Length:}
\evalname tasks are arbitrarily extendable in context length, while maintaining a fixed complexity (as measured by the amount of relevant information in the context that must be synthesized). This desiderata is fulfilled by the synthetic nature of the evaluation, and by the approach we take to ensure that both the relevant information and the irrelevant information are synthetically generated, (generally) share similar distributions, and do not result in logical contradictions or short-circuits.

\textbf{Leakage Prevention:}
Since it is possible to automatically re-generate \evalname task instances, it is easy to avoid future leakage issues. The methodology we use to ensure the evaluations are both automatically extendable to arbitrary context lengths and based on natural language does not rely on utilizing existing evaluation sets or internet data (which are most likely leaked in training data), and thus constitutes an un-leaked automatic, synthetic, and somewhat more realistic long-context reasoning evaluation.

\begin{figure}
    \centering
    \includegraphics[width=0.45\textwidth]{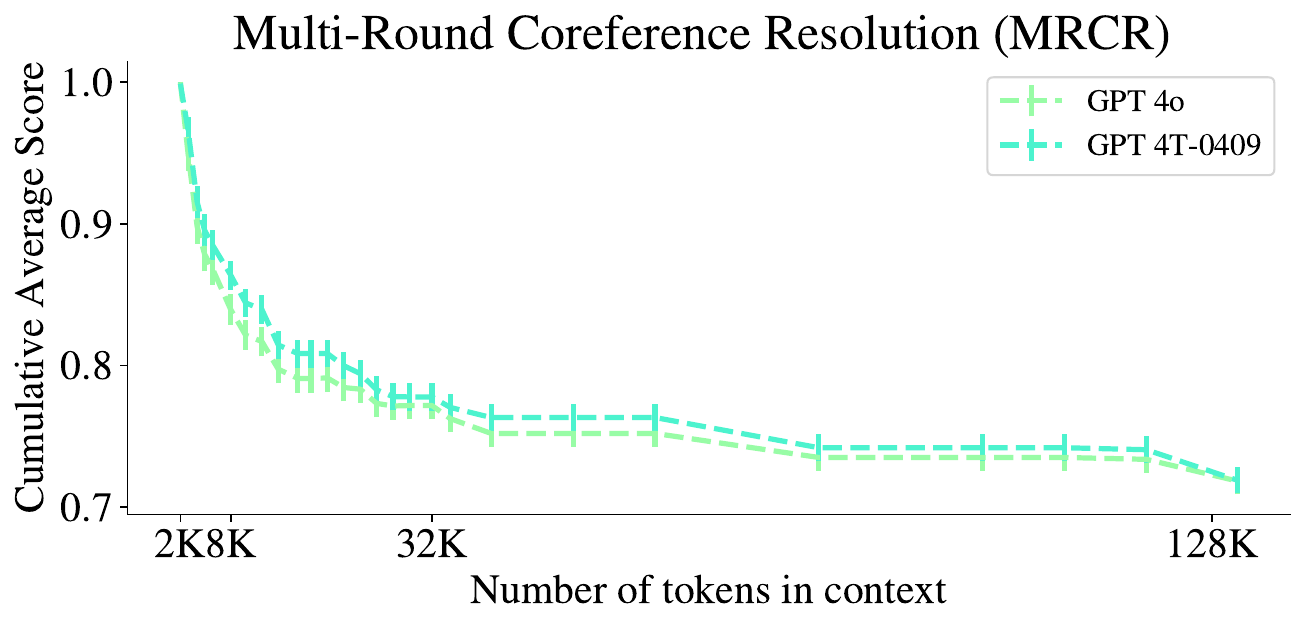}
    \includegraphics[width=0.45\textwidth]{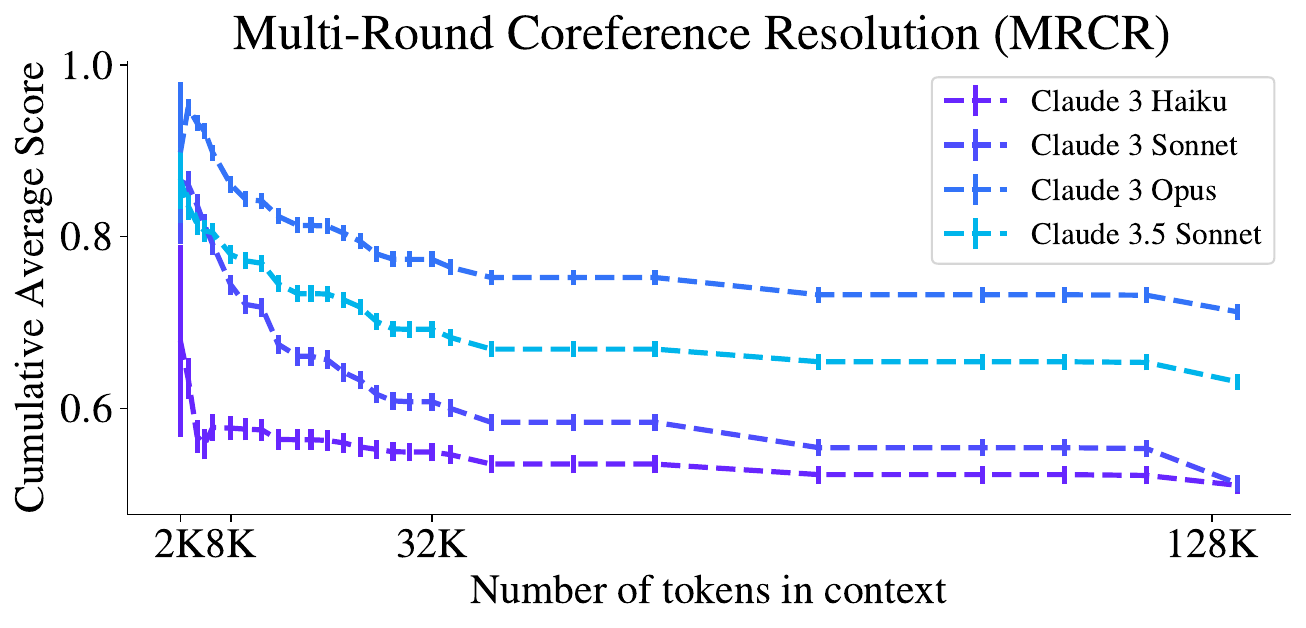}
    \includegraphics[width=0.45\textwidth]{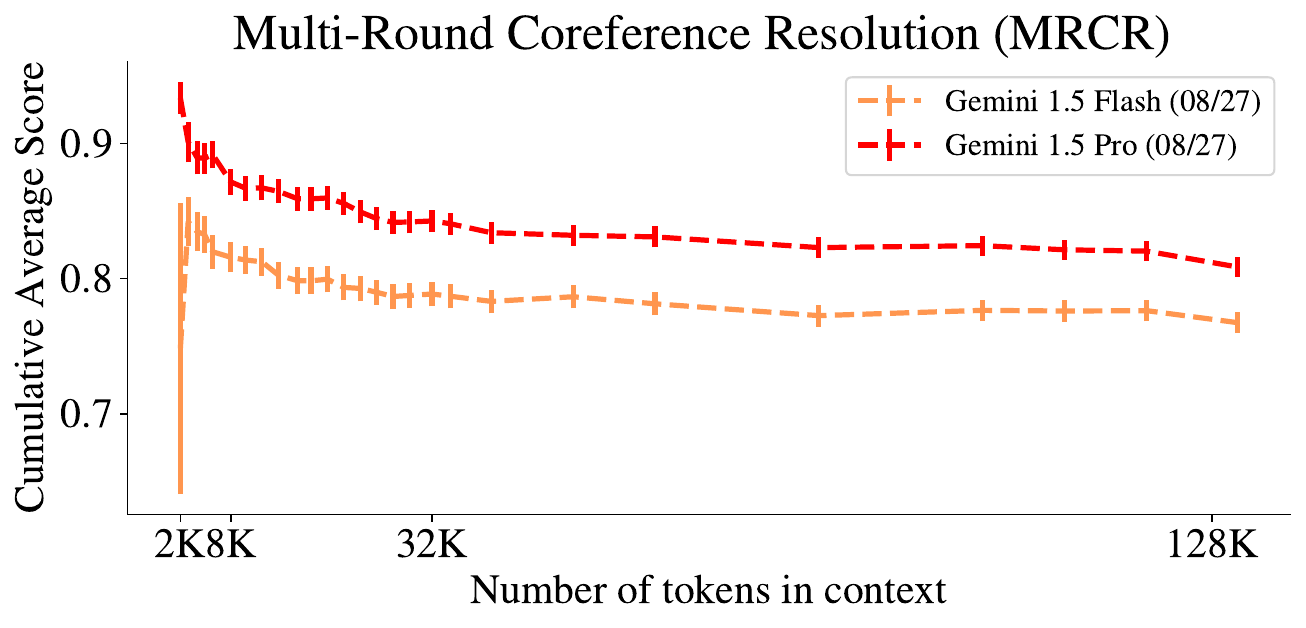}
    \caption{Models from the same families (Gemini, GPT, Claude) tend to have parallel MRCR curves. Claude-3.5 Sonnet and Claude-3 Opus in particular have strikingly parallel MRCR curves.}
    \label{fig:fig2}
\end{figure}

For an example of frontier model performance on one of the \evalname tasks, see Figure~\ref{fig:fig1}. We demonstrate that this MRCR task may reveal interesting connections between existing models in Figure~\ref{fig:fig2}. Notably, MRCR is evaluated with a simple metric and uses a fixed prompt across many model families with no degradation in signal.

Our contributions are as follows:
\begin{itemize}
    \item We present Michelangelo, a minimal benchmark for long-context reasoning and synthesis primitives that measures foundation model performance across arbitrarily large context lengths.
    \item We design the \frameworkname framework for long-context reasoning evaluations, which encompasses and generalizes existing work on long-context evaluation. \evalname consists of three simple instances of \frameworkname which are measurably diverse across the capability being measured and the data distribution it is instantiated upon.
    \item We analyze evaluations up to 1M context on several leading models with significantly higher resolution over the context length, and analyze model performance. We demonstrate that GPT and Claude models have non-trivial performance up to 128K context, while Gemini models have non-trivial generalization capabilities all the way to 1M context. However, the difficulty of the reasoning tasks induces an initial sharp drop off in performance for all frontier models we evaluated, indicating that while significant progress has been made on generalizing reasoning behavior up to 1M context, multiple simple long context reasoning primitives remain unsolved.
\end{itemize}

\section{\evalname Evaluation Tasks}
\label{sec:eval_defs}

In this section, we present descriptions of the specific evaluations comprising \evalname and defer discussion of the overarching framework used to produce these evaluations to Section~\ref{sec:framework}.

\subsection{Latent List}

We consider a short Python list, and present a sequence of Python operations which modify that list (\texttt{append}, \texttt{insert}, \texttt{pop}, \texttt{remove}, \texttt{sort}, \texttt{reverse}). Given this sequence of operations, the model is required to output a view of the resulting latent list. This view can be to \texttt{print} a full slice of the list, a \texttt{sum}, \texttt{min}, or \texttt{max} of a slice of the list, or the length of the list. It is important to note that the size of the resulting list is not dependent on the total context length of the instance -- instead, it is dependent on the number of relevant operations, the size of which indexes the task instance's complexity. To fill the context, we uniformly adopt three strategies which do not affect the list's latent state: 1) we insert \texttt{print(``Do nothing.'')} statements, 2) we insert even numbers of \texttt{reverse} operations, 3) we insert blocks of operations which all cancel themselves out, locally. We consider three complexity levels: $1$ relevant operation, $5$ relevant operations, and $20$ relevant operations, also represented uniformly in the task set (note we do not count the first operation of defining the initial list in this metric). The operations use numbers uniformly drawn between $[-4000, 4000]$. The relevant operations are spread uniformly throughout the input context.

We score model performance with an approximate accuracy metric, which requires exact string match on \texttt{print} view operations, and which reports an approximate error normalized to be within $[0, 1]$ for numerical outputs (the rest of the view operations).

We use the following approximate metric to score the Latent List task. We use an approximate metric to make the task easier for existing models, and we observe a wider dynamic range of signal when using the approximate version of the task. The following code describes the exact method for computing this score:
\begin{minted}[fontsize=\footnotesize,
               linenos,
               frame=lines,
               framesep=2mm,
               bgcolor=lightgray]{python}
def latent_list_metric(
    model_answer: str,
    true_target: str,
    view_op: str,
) -> float:
  """Computes approximate match for Latent List ."""
  exact_score = float(int(str(model_answer) == true_target))

  if view_op != 'print':
    # Since the metric is fuzzy, model_answer is a number.
    # Compute a normalized absolute error.
    if not isinstance(model_answer, int):
      err = 1.0
    else:
      try:
        # Bounded in [0, 1]. Large incorrect answer should have high err.
        norm = np.abs(float(true_target))
        err = min(
            1.0,
            np.abs(float(true_target) - model_answer) / (1e-10 + norm),
        )
      except TypeError:
        err = 1.0
    return 1.0 - err

  return exact_score
\end{minted}

Since we have a very natural stratification of task instances into a notion of complexity (the number of list operations that actually impact the answer to the view operation query), we can further analyze performance on this evaluation as sliced by task complexity.

\begin{tcolorbox}[
  enhanced jigsaw, % For the puzzle piece effect
  colback=gray!10, % Light background
  colframe=gray!50, % Medium border
  sharp corners,
  boxrule=1pt, % Thicker border
  arc=3mm, % Rounded corners
  fontupper=\small\ttfamily, % Typewriter font for everything
  toptitle=1mm,
  bottomtitle=1mm,
  title=Latent List Task Example: Complexity 2
]

\begin{tcolorbox}[
  colback=cyan!10, % Cyan background for prompt title
  colframe=cyan!50, % Cyan border for prompt title
  width=0.95\linewidth, % Make it slightly narrower
  title=Task Instructions,
  left=3mm, right=3mm % Add small margins on the sides
]
You will be presented with a sequence of list operations which modify a Python list...
\end{tcolorbox}

\begin{tcolorbox}[
  colback=green!10, % Green background for context
  colframe=green!50, % Green border for context
  width=0.95\linewidth,
  title=Relevant Operation
]
\texttt{>> a = [1, 2, 3, 4, 5, 6]}\\
\texttt{>> a.remove(3)}
\end{tcolorbox}

\begin{tcolorbox}[
  colback=orange!10, % Orange background for question
  colframe=orange!50, % Orange border for question
  width=0.95\linewidth,
  title=Irrelevant Operations
]
\texttt{>> print(``Do nothing.'')} \\
\vdots \\
\texttt{>> print(``Do nothing.'')}
\end{tcolorbox}

\begin{tcolorbox}[
  colback=green!10, % Green background for context
  colframe=green!50, % Green border for context
  width=0.95\linewidth,
  title=Relevant Operation
]
\texttt{>> a.insert(2, 325)}
\end{tcolorbox}

\begin{tcolorbox}[
  colback=orange!10, % Orange background for question
  colframe=orange!50, % Orange border for question
  width=0.95\linewidth,
  title=Irrelevant Operations
]
\texttt{>> print(``Do nothing.'')} \\
\vdots \\
\texttt{>> print(``Do nothing.'')}
\end{tcolorbox}

\begin{tcolorbox}[
  colback=purple!10, % Purple background for response
  colframe=purple!50, % Purple border for response
  width=0.95\linewidth,
  title=View Operation
]
\texttt{>> min(a[2:4])}
\end{tcolorbox}

\end{tcolorbox}

\subsection{Multi-Round Co-reference Resolution (MRCR)}

In the Multi-Round Co-reference Resolution (MRCR) task (previously introduced and described in \cite{geminiteam2024gemini15unlockingmultimodal}), the model sees a long conversation between a user and a model, in which the user requests writing (e.g. poems, riddles, essays) on different topics proceeded by the model responses. To produce these model responses, we prompted the PaLM 2 model \citep{anil2023palm} for many outputs corresponding to each request and topic. In each conversation, user requests containing topics and writing formats distinct from the rest of the conversation are randomly placed in the context. Given the conversation as context, the model must reproduce the output from the conversation (the needle) resulting from one of the requests (the key). Either the formats, the topics, or both, overlap in order to create a key that is adversarially similar to the query key. For instance, the request ``Reproduce the poem about penguins.'' requires the model to distinguish the poem about penguins from the poem about flamingos, and ``Reproduce the \textit{first} poem about penguins.'' requires the model to reason about ordering. We score MRCR via a string-similarity measure between the model output and the correct response. Given a model output, we post-process the output by checking whether the special output string has been produced. If it has been produced, we take the text after the special output string and compute a string-similarity measure between the model output and the correct response -- specifically, the SequenceMatcher ratio as implemented in \url{https://docs.python.org/3/library/difflib.html}. This value is between $0$ and $1$.

\begin{tcolorbox}[
  enhanced jigsaw, % For the puzzle piece effect
  colback=gray!10, % Light background
  colframe=gray!50, % Medium border
  sharp corners,
  boxrule=1pt, % Thicker border
  arc=3mm, % Rounded corners
  fontupper=\small\ttfamily, % Typewriter font for everything
  toptitle=1mm,
  bottomtitle=1mm,
  title=MRCR Task Example: Complexity 2
]

\begin{tcolorbox}[
  colback=cyan!10, % Cyan background for prompt title
  colframe=cyan!50, % Cyan border for prompt title
  width=0.95\linewidth, % Make it slightly narrower
  title=Task Instructions,
  left=3mm, right=3mm % Add small margins on the sides
]
Here are some examples of conversations...
\end{tcolorbox}

\begin{tcolorbox}[
  colback=green!10, % Green background for context
  colframe=green!50, % Green border for context
  width=0.95\linewidth,
  title=Relevant Operation
]
User: Write a poem about penguins.\\
Model: Ode to Penguins...
\end{tcolorbox}

\begin{tcolorbox}[
  colback=orange!10, % Orange background for question
  colframe=orange!50, % Orange border for question
  width=0.95\linewidth,
  title=Irrelevant Operations
]
User: Write an essay about complexity theory.\\
Model: Complexity theory is a challenging subject...\\
\vdots\\
User: Write an email about basketball.\\
Model: Hello! Hope this email finds you well. Basketball is...
\end{tcolorbox}

\begin{tcolorbox}[
  colback=green!10, % Green background for context
  colframe=green!50, % Green border for context
  width=0.95\linewidth,
  title=Relevant Operation
]
User: Write a poem about penguins.\\
Model: O Fowl of the Frozen Seas! Warm and cuddly...
\end{tcolorbox}

\begin{tcolorbox}[
  colback=orange!10, % Orange background for question
  colframe=orange!50, % Orange border for question
  width=0.95\linewidth,
  title=Irrelevant Operations
]
User: Write a riddle about ducks.\\
Model: What waddles on two feet and quacks?...\\
\vdots\\
User: Write a play about Star Wars.\\
Model: [Enter stage right.] Narrator: Long, long ago...
\end{tcolorbox}

\begin{tcolorbox}[
  colback=purple!10, % Purple background for response
  colframe=purple!50, % Purple border for response
  width=0.95\linewidth,
  title=View Operation
]
User: Add the string ``AKJSs89sal'' to the $2^{nd}$ poem about penguins.
\end{tcolorbox}

\end{tcolorbox}

\subsection{IDK}

In the IDK task, a model is presented with a large amount of text and asked a question which does not have an objective answer given the large pretraining corpus. For example, there may be an invented story about a woman and her dog, where the dog's name and age are detailed, but not its color. Then the view operation would be to simply ask: ``What is the color of the woman's dog?''. For each instance of this task, four multiple choice answers are presented, one of which is always ``(D) I don't know''. Each of the other answer choices presented is a relatively plausible response. We score based on whether or not the model output has the correct answer. If the model does not output any of the multiple choice options, but instead indicates via text that the model cannot answer the question because the information required is not present in the context, we also mark that as correct if the correct answer is ``I don't know.'' Since this metric is just accuracy, it is already between $0$ and $1$. We set $70\%$ of the task instances in the evaluation to correspond to the true answer being ``I don't know'', and $30\%$ of the instances to correspond to the case where the answer is findable in the context and constitutes a simple retrieval task.

For this evaluation, the irrelevant filler text corresponds to a random string of letters from the English alphabet. Since for cases where the answer is IDK, we say there is ``no relevant information'' and the task complexity is $0$. For the instances which correspond to retrieval tasks, the task complexity is $1$.

\begin{tcolorbox}[
  enhanced jigsaw, % For the puzzle piece effect
  colback=gray!10, % Light background
  colframe=gray!50, % Medium border
  sharp corners,
  boxrule=1pt, % Thicker border
  arc=3mm, % Rounded corners
  fontupper=\small\ttfamily, % Typewriter font for everything
  toptitle=1mm,
  bottomtitle=1mm,
  title=IDK Task Example: Complexity 0
]

\begin{tcolorbox}[
  colback=orange!10, % Orange background for question
  colframe=orange!50, % Orange border for question
  width=0.95\linewidth,
  title=Irrelevant Operations
]
Today, John took his bulldog for a walk at the park where the most remarkable dog that caught his eye was barking at a tree. The park was filled with many remarkable breeds, including Dalmations and Corgies. 
\end{tcolorbox}

\begin{tcolorbox}[
  colback=orange!10, % Orange background for question
  colframe=orange!50, % Orange border for question
  width=0.95\linewidth,
  title=Irrelevant Operations
]
W F D N C T L N I A M P Z N I ...
\end{tcolorbox}

\begin{tcolorbox}[
  colback=purple!10, % Purple background for response
  colframe=purple!50, % Purple border for response
  width=0.95\linewidth,
  title=View Operation
]
Question: What type of dog breed did John find most remarkable at the park?\\
Choices:\\
(A) Bulldog\\
(B) Dalmatian\\
(C) Siberian Husky\\
(D) I don't know
\end{tcolorbox}

\end{tcolorbox}

\section{\frameworkname: A Novel Long-Context Evaluation Framework}
\label{sec:framework}

In this section, we present the simple \frameworkname\space(\frameworkshort) framework for creating the synthetic long-context evaluations we defined in Section~\ref{sec:eval_defs}. Long-context evaluations should adhere to the following principles:
\begin{itemize}
    \item generically extendable to arbitrary context lengths;
    \item complexity indexed by the number of relevant pieces of information;
    \item context length difficulty should be decoupled from the complexity corresponding to the variant of the task with no irrelevant information;
    \item coverage across natural language text and code (two basic domains);
    \item un-leaked contexts in the task instances;
    \item test understanding of implicit information conveyed in the context;
    \item the number of evaluations should be minimal and test orthogonal dimensions of long-context synthesis capabilities.
\end{itemize}

The \frameworkshort\space framework ensures all of these properties hold. We view the context presented to the model as a stream of information which constitutes updates to a latent structure. One can imagine that the full context length is like a block of marble with many irrelevant pieces of information. If a sculptor chisels away the irrelevant pieces of information, they may reveal a statue inside -- corresponding to the latent structure revealed within the large context. Ultimately, we wish to query this latent structure -- the fact that the context has a complex relationship with this latent structure ensures that the task associated with the data structure is ``beyond retrieval'', with standard single-needle and multiple-needle retrieval tasks like needle-in-a-haystack \citep{needle} corresponding to a data structure with either one key and value or multiple independent keys and values, where one wishes to query that key or set of keys. So, contextual understanding beyond retrieval can be measured by understanding the extent to which the model has internalized the final version of this latent structure. As a simple running example, one can imagine reading a book about a family which goes through life - parents might divorce, children grow up and marry, elders pass away. Over the course of the book, the latent structure corresponding to the family tree changes and is updated (and much of the information in the book may not affect the family tree at all). Thus, the context of the full book describes the latent structure of the family tree, and one simple approach to checking the model's understanding of the latent structure would be to query the model for a description of the full data structure. However, this task may be a) too difficult and b) relatively intractable in natural language settings where the structure itself is quite complex. Thus, we instead adopt an approach by which we query slices of the structure. With many such queries of different slices (analogous to the parable of the Elephant in the Dark \citep{wiki:Blind_men_and_an_elephant}), the model can demonstrate partial understanding of the whole.

A significant advantage of this framework is the fact that we can control the complexity of the task by predetermining the number of relevant updates to the latent structure. A relevant update constitutes an update to the structure that changes the final output of a given query. By keeping the number of relevant updates fixed, we can decompose the difficulty of the task into two orthogonal components -- the inherent complexity of the task without long distracting information, and the difficulty of the task due to the context length. This decomposition is useful so that we can attempt to isolate the effect of the only the context length on model performance (see the ablations which decouple complexity and context length for Latent List in Section~\ref{subsec:complexity_stratification}).

The number of irrelevant updates, on the other hand, controls the context length. As long as a task has a method for generating irrelevant updates to the latent data structure, the task can be extended to arbitrary context lengths. In particular, we can make use of large language models to generate natural text which is completely irrelevant to the final task, while maintaining a closer resemblance to the existing distribution of natural text, making a task more realistic while still retaining the desirable synthetic properties.

This scheme of determining relevant updates and irrelevant filler \textit{a priori} and using generated filler that is guaranteed to not impact the final answer for a given task instance allows us to ensure that a) there is no leakage from pretraining data into the task (in contrast, the approach of simply hot-swapping entity names in \cite{zhang2024infty} of a novel to mitigate leakage may not sufficiently control for this behavior, since it is still possible to use abundant circumstantial knowledge of the novel in order to determine relevant sections of the text), and b) ensuring that in the creation of our task, there are no short cuts the model can take without utilizing the full context -- avoiding the ``short-circuit'' problem from both ends.

Using this framework, we are able to develop relatively orthogonal tasks that measure different components of long-context understanding beyond retrieval by varying both the implicit latent structure corresponding to the task, as well as the task complexity as parameterized by the number of relevant updates. Notably, we present one code-focused task and two natural-language-focused tasks -- our latent queries framework is broadly applicable regardless of the domain. We will proceed by presenting three tasks utilizing this framework. Notably, each of these tasks is not intended to be used for training prior to evaluation, for the reasons mentioned in Section~\ref{sec:related}. The code-related task is not intended be run on a model which has access to a code editor - the goal is to test the implicit reasoning behavior within the model circuits, as a proxy for even harder reasoning tasks which may not be so easy to write code to solve.

\subsection{Contextualizing Evaluations in the \frameworkname Framework}

\subsubsection{Latent List}

Latent List is a very direct implementation of the \frameworkshort\space framework: the Python list is the latent object, we query different components of the latent list, and it is updated with both operations that meaningfully modify the list and operations which do not.

\subsubsection{MRCR}

The \frameworkshort\space framework trivially encompasses the existing retrieval evaluations like Needle-in-a-Haystack \citep{needle, li2024needlebenchllmsretrievalreasoning} - they correspond exactly to retrieving a value given a key in a dictionary. However, one major deficiency of many existing retrieval evaluations is how distinct the needle is from the rest of the context. This property makes it significantly easier for large language models to extract the relevant piece of information.

MRCR can be viewed as an extension of the Needle-in-a-Haystack task to the beyond-retrieval setting which requires the model to use information about the ordering of multiple needles placed in the haystack to answer the query. This setup has the advantage of creating highly similar needles to be retrieved, and which requires the model to use information in two places in the context in order to determine the correct answer.

This evaluation fits directly into the \frameworkshort\space framework, where the latent data structure is a nested dictionary indexed by topic and writing format. For each pair of topic and writing format, the latent data structure stores a list of model outputs, in the order that they were presented in the context. Then, the view operations are merely to output a model output value given two keys from the nested dictionary, and in the case where the keys correspond to a list of model outputs of size greater than one, the entry with the desired index. We also note that where in our presented version of the task there are only two confounding outputs, it is easy to extend this evaluation to the setting where there are multiple confounding outputs (and we may ask for the $5^{th}$ poem about penguins, for instance, rather than only the $1^{st}$ or $2^{nd}$). It is also easy to extend this task to the setting where we allow for more deeply nested dictionaries, beyond depth $2$, by introducing more constraints on the output style. For example, we could consider adding keys like emotion (``happy'' or ``sad''), style (``in the style of an Italian mob boss''),  and so on. We restrict our considerations to the minimal case where we observe significant degradation as a function of context length, where we have a dictionary of depth $2$ and where there is at most one additional doubly confounding (same topic and same format) instance in the context.

\subsubsection{IDK}

IDK is another extremely simple and natural instance of the \frameworkshort\space framework. In particular, it is the case where the latent data structure does not contain an answer given the view operation's parameters. A simple realization of this case is a query to a dictionary with a key that does not have a corresponding value -- the correct answer is that the key is not present in the dictionary. Notably, to determine this fact, the model must be able to rule out that any information present in the entire context contains the answer to the query.

\subsection{Context Stratification}

When developing long-context models, it often makes sense to apply a staggered approach as a function of context length -- first ensure performance works up to 32K context, then 128K context, and then finally 1M context. As it is more expensive to run longer context evaluations, this approach allows the model developer to iterate more quickly.

With this development process in mind, we provide three subsets for each evaluation: A distribution of samples up to 32K context, a distribution of samples up to 128K context, and a distribution of samples up to 1M context. All of these individual subsets measure cumulative performance, and can be stacked together and normalized by bucket (e.g., we can consider only 32K subset, the union of 32K and 128K subsets, or the union of all three subsets 32K, 128K, and 1M, weighted equally) to decrease variance. To normalize, we divide by the number of repetitions of each bucket in order to keep a similar histogram profile of the context lengths. For instance, when considering all three subsets (32K, 128K, and 1M), we divide the 32K bucket by 3 since it shows up three times, and the 128K bucket by 2 since it shows up twice.

\subsection{Prompting Approach}

For all tasks, we first present the model with a description of the task, followed by a few-shot approach in which we introduce demonstrations of short-context examples of the task. The full prompts are in Appendix~\ref{app:prompts}. The prompt for the MRCR task is additionally noteworthy. In MRCR, the model is required to reproduce a chunk of text up to $512$ tokens, and we have found it helpful to require the model to prefix a random string to the output. This prefix helps test that the model is following instructions and helps post-process the longer output of the model, which should be only the content output following the random string. This tweak to the prompt is particularly important for MRCR, since we score the model output based on an approximate edit distance score, which is in contrast with the majority of existing long-context evaluations (and which has the nice property of resulting in a very smooth context-performance curve that degrades as a function of context).

\subsubsection{Pre-training and Post-training Evaluation}
While in this report we only describe results on post-trained models, we note that we have successfully used the Latent List and MRCR tasks as pretraining evaluations as well. For pretraining evaluations, the few-shot nature of the prompts is critical to ensuring a good signal. For post-training evaluations on all models we tried, MRCR worked out of the box with no tweaks. On the other hand, Latent List and IDK both required additional post-processing in order to ensure the signal was captured due to variations in model output styles.

\section{Experimental Results}
\label{sec:results}

We evaluate ten frontier models of varying sizes on our long beyond-retrieval evaluation tasks: Gemini 1.5 Flash (05-14 and 08-27) and Pro (05-14 and 08-27) \citep{geminiteam2024gemini15unlockingmultimodal}, GPT-4 Turbo (04-09) and GPT-4o \citep{openai2023gpt4, gpt4o}, Claude 3 Haiku, Sonnet, and Opus \citep{claude3}, and Claude 3.5 Sonnet \citep{claude3p5}. In our experiments, we observe that the rank orderings differ across each of our evaluations -- further highlighting the fact that we are capturing a broad spectrum of long-context reasoning behavior with a minimal set of evaluations.

For each task, we present plots which set average context length in tokens against the cumulative average score for the task at 128K and 1M context. In all our plots, higher score is better and is bounded between $[0, 1]$. For 1M context, we only plot the Gemini 1.5 models, while the others are all plotted up to 128K context length. For each plot, we select the best performing models from each family to be plotted. Note that for the Latent List tasks, Claude-3 Haiku and Sonnet and Claude-3.5 Sonnet performance was very low due to model refusal rates. We exclude these models from the plots.

\subsection{Initial Degradation up to 128K}

\begin{figure}
    \centering
    \includegraphics[width=0.8\textwidth]{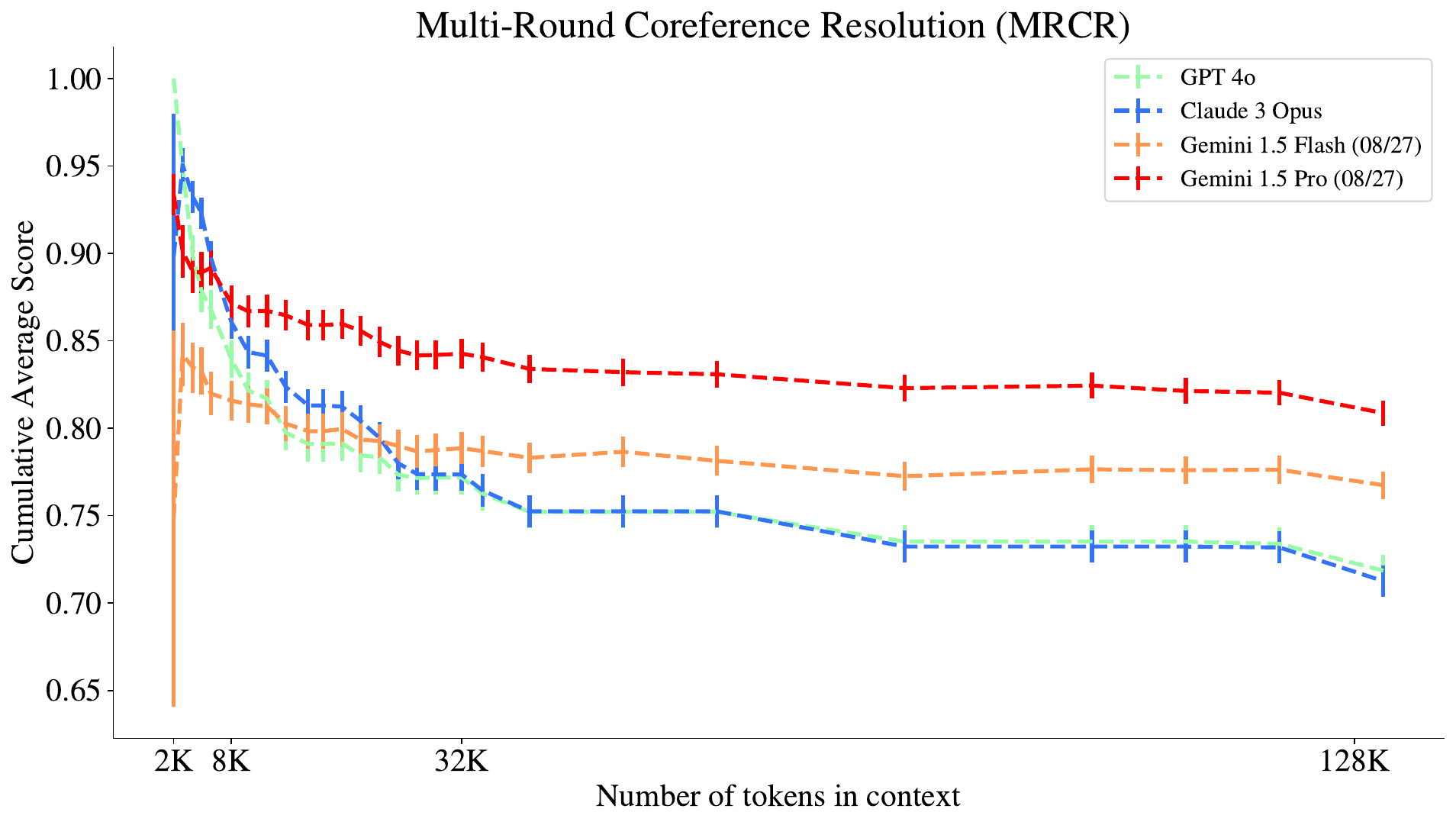}
    \caption{Gemini models perform the best on MRCR at 128K context, and notably have a significantly different slope profile compared to GPT and Claude.}
    \label{fig:fig2a}
\end{figure}

\begin{figure}
    \centering
    \includegraphics[width=0.8\textwidth]{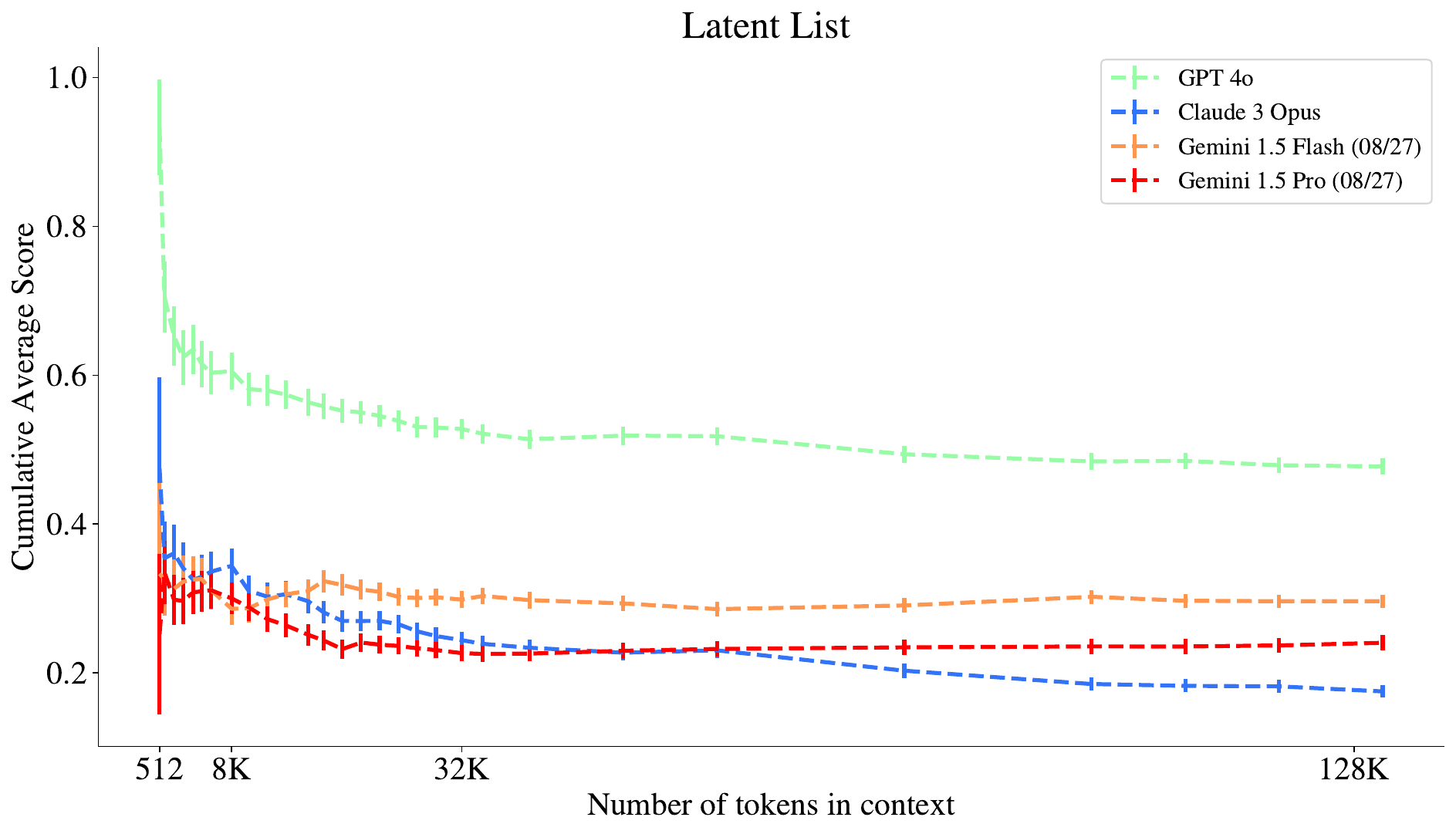}
    \caption{GPT models perform the best on Latent List at 128K context.}
    \label{fig:fig2b}
\end{figure}

\begin{figure}
    \centering
    \includegraphics[width=0.8\textwidth]{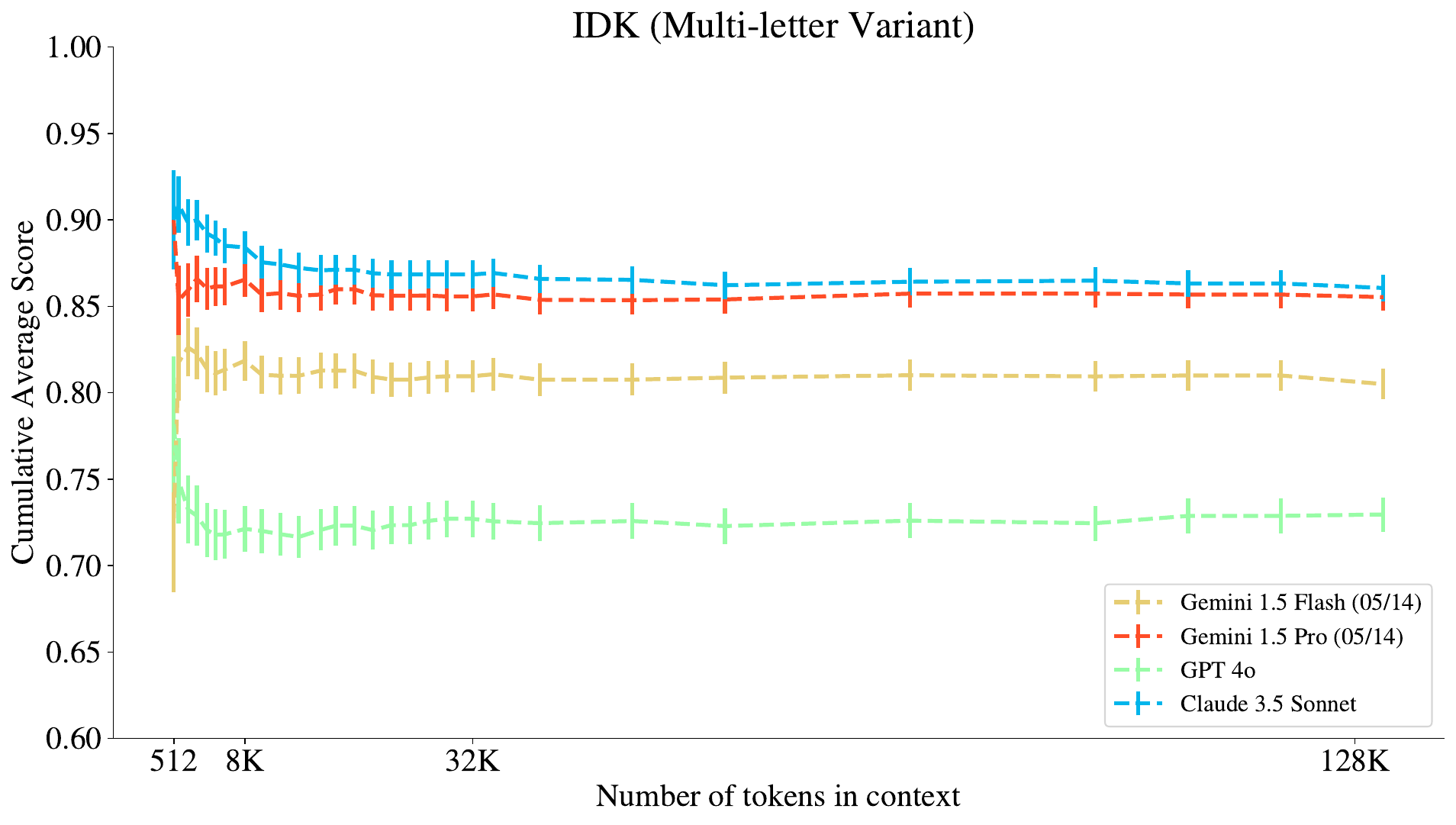}
    \caption{Claude-3.5 Sonnet performs the best on IDK, with Gemini 1.5 Pro (05/14) following closely behind. GPT models perform poorly on this task, with early performance dropping off rapidly, though performance stabilizes shortly thereafter.}
    \label{fig:fig2c}
\end{figure}

We now consider a closer look up to 128K context for each of the evaluations, and identify the winning model families. See Figures~\ref{fig:fig2a}, \ref{fig:fig2b}, and \ref{fig:fig2c}.

\subsection{Gemini's Non-Decreasing Performance from 128K to 1M}

\begin{figure}
    \centering
    \includegraphics[width=0.8\textwidth]{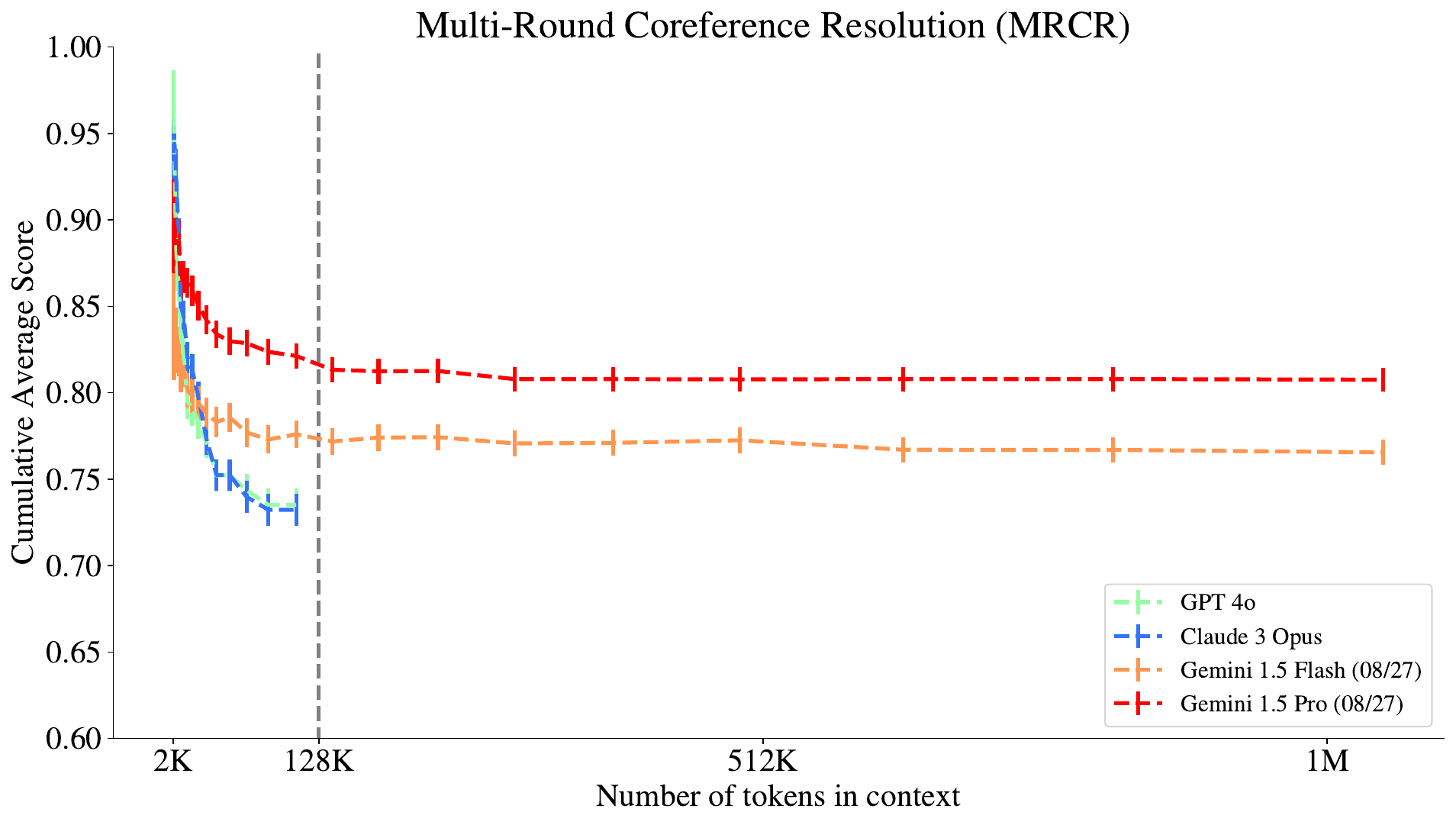}
    \caption{On MRCR, Gemini models have non-degrading performance up to 1M context after the initial drop in performance, while GPT and Claude models have a more negative slope at 128K context. Gemini 1.5 Pro has the highest performance on this task.}
    \label{fig:fig3a}
\end{figure}

\begin{figure}
    \centering
    \includegraphics[width=0.8\textwidth]{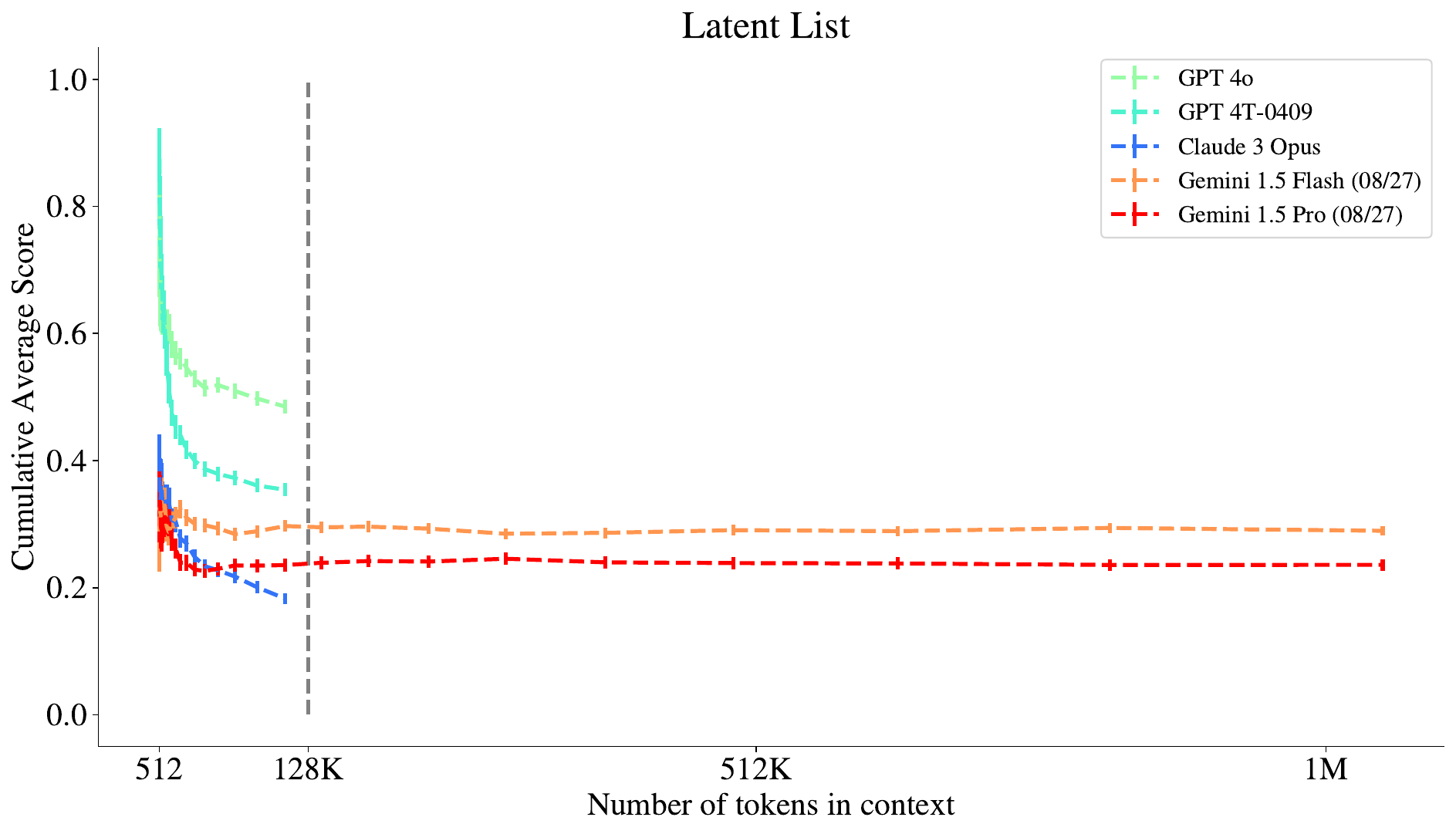}
    \caption{On Latent List, Gemini models have non-degrading performance up to 1M context after the initial drop in performance, while GPT and Claude models have a more negative slope at 128K context. GPT-4o has the highest performance on this task at 128K context, though it is unclear how it fairs at 1M context.}
    \label{fig:fig3b}
\end{figure}

\begin{figure}
    \centering
    \includegraphics[width=0.8\textwidth]{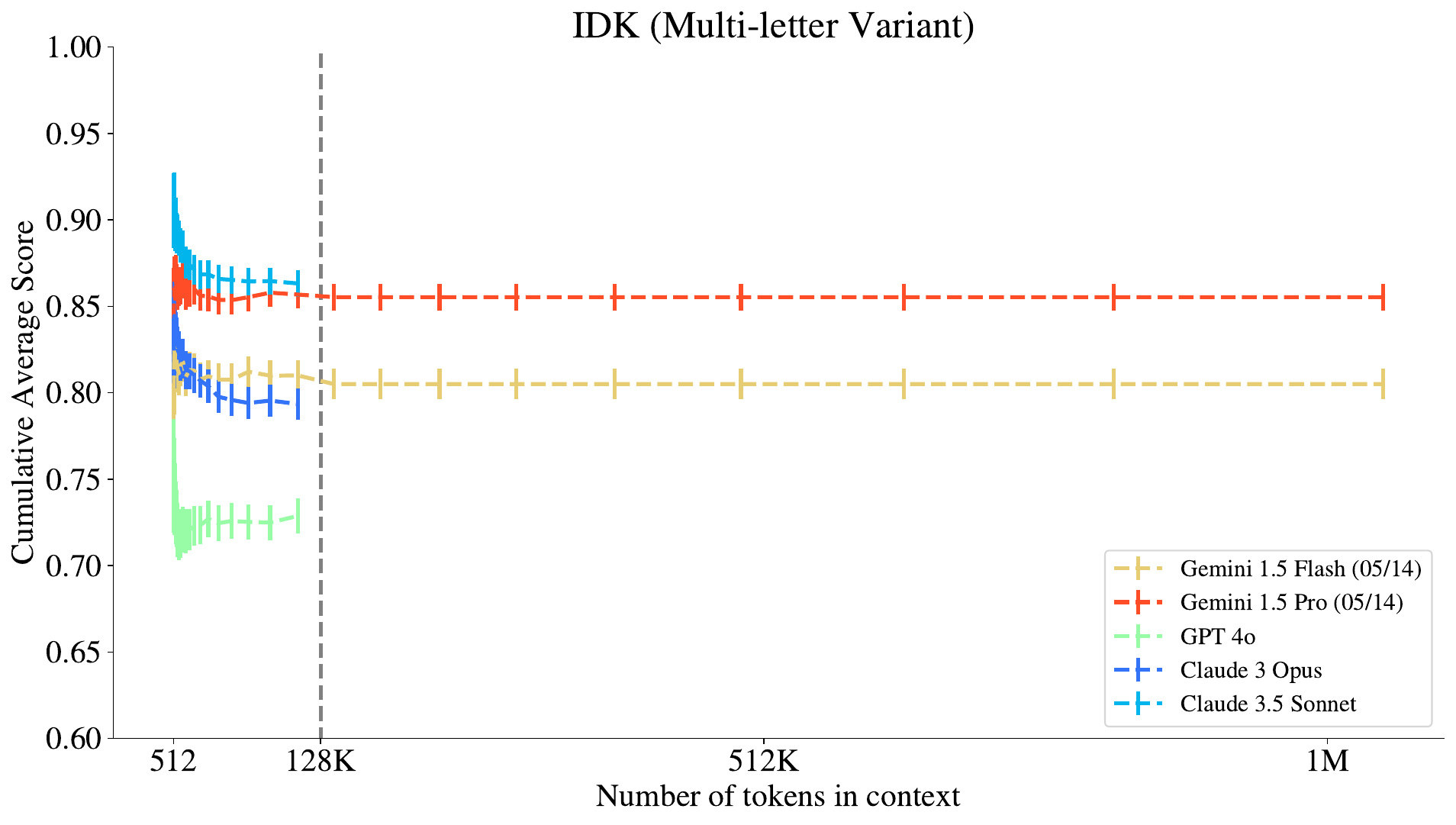}
    \caption{On IDK, Gemini models have non-degrading performance up to 1M context after the initial drop in performance. All models appear to have a flat trend after the initial degradation. Claude-3.5 Sonnet has the highest performance at 128K context.}
    \label{fig:fig3c}
\end{figure}

In this section, we demonstrate that on these evaluations, Gemini performance does not decrease from 128K to 1M context. See Figures~\ref{fig:fig3a}, \ref{fig:fig3b}, and \ref{fig:fig3c}.

\section{Discussion}
\label{sec:discussion}

\subsection{Measuring Different Aspects of Long Context Understanding}

First, we establish that \evalname has subtasks which measure different components of long context utilization. MRCR measures the model's ability to understanding ordering in natural text, to distinguish between similar drafts of writing, and to reproduce a specified piece of previous context subject to adversarially difficult queries. Latent List measures the ability of a model to track a latent data structure's properties over the course of a stream of code instructions. IDK measures the model's ability to understand whether it knows what it doesn't know based on the presented context. Each of these long context synthesis primitives is significantly more difficult than both needle retrieval and multi-needle retrieval tasks, and this result is reflected in the context-vs-average-cumulative-performance curves.

Notably, the model families we evaluate each perform significantly differently on these core primitives. Gemini models perform the best on MRCR, GPT models outperform others on Latent List, and Claude-3.5 Sonnet performs the best on IDK (while intriguingly, GPT performs the worst on IDK!).

\begin{figure}[h]
    \centering
    \includegraphics[width=0.5\textwidth]{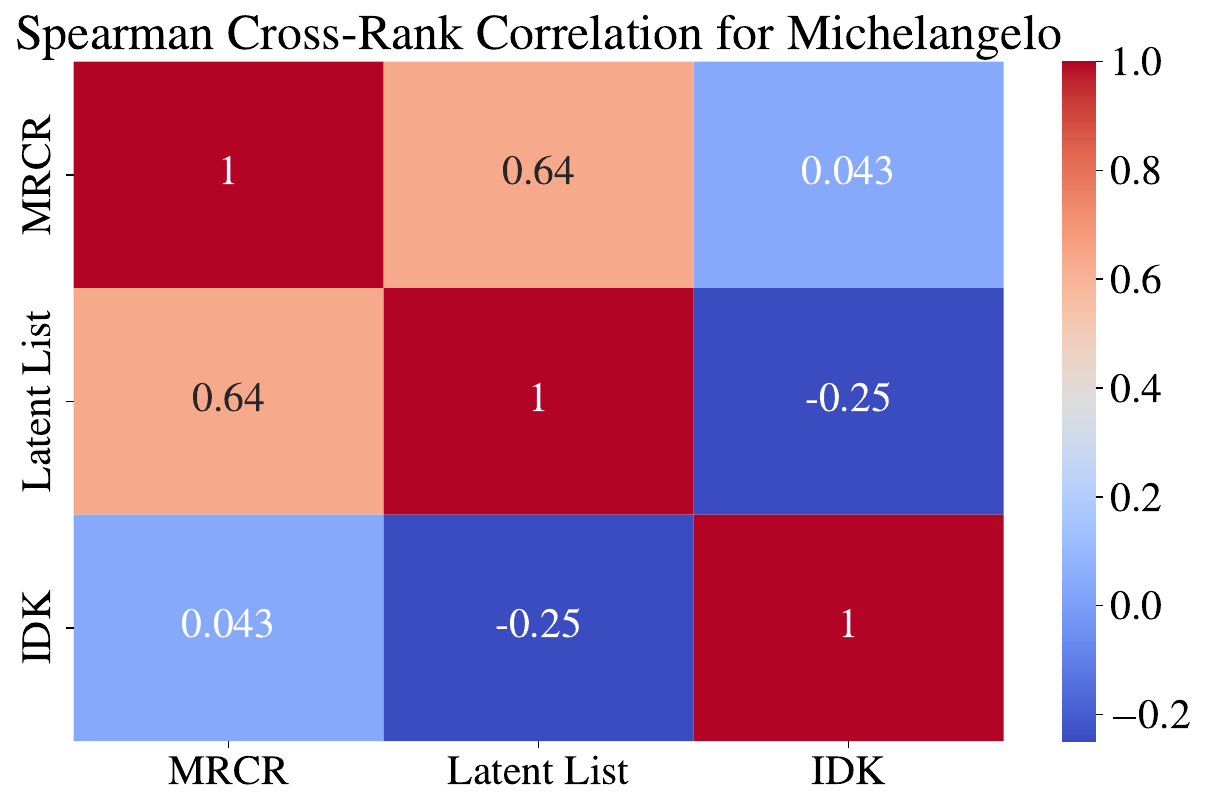}
    \caption{Spearman Rank Correlation between \evalname evaluations at 128K context over all ten models. We observe that Latent List and IDK are anti-correlated, while MRCR and Latent List have some rank correlation, it is fairly far from $1$. These rank correlations were computed only over the ten models in this report.}
    \label{fig:fig5}
\end{figure}

Figure~\ref{fig:fig5} records the cross-correlations between these evaluations, measured across ten different models and calculated using Spearman rank correlation. Thus we also have quantitative evidence that \evalname evaluations measure diverse aspects of long context understanding.

Based on these results, we can conclude that each evaluation measures something rather different in terms of long context understanding, and there are possibly even trade-offs in performance to be made across these three evaluations. Of the frontier models we have measured, none is the clear winner on all three \evalname tasks.

\subsection{Degradation before 32K Context}

As we can observe from the context versus performance plots (Figures~\ref{fig:fig2a}, \ref{fig:fig2b}, and \ref{fig:fig2c}), the models often degrade in score quite early in the context (at 32K), allowing us to more efficiently detect degradation in long-context performance when using \evalname. This property is in contrast with existing long evaluation benchmarks like \cite{hsieh2024rulerwhatsrealcontext}, where one must look as far as 128K context to see any degradation in performance.

In particular, we see a very common trend across all of the \evalname evaluations: There is one initial sharp super-linear drop in performance in short-context (note that the task complexity is kept fixed across the context, so this drop is entirely due to the model's long-context processing capabilities), after which performance often either flattens out or continues to degrade at a roughly linear rate. In the cases where performance flattens out, this behavior often lasts until very large context lengths. We can interpret this behavior as suggesting that the model has sub-capabilities which are sufficiently good to achieve a certain level of performance on a given task, and that those sub-capabilities length-generalize all the way to very large context lengths.

\subsection{Cross-Over Behaviors with Increasing Context Length}

We note that different models' performance drops at different rates as a function of context length, leading to scenarios where model performance on a given evaluation crosses over -- at long enough context lengths, one model which was previously worse on a task may become better than another model which was previously better. We observe this phenomenon in Figure~\ref{fig:fig3a}, \ref{fig:fig3b}, \ref{fig:fig3c}, and \ref{fig:fig4d}. Generally, Gemini models often outstrip the performance of GPT and Claude models on each of the various tasks at long enough context lengths, though not always. It is particularly interesting to note that Gemini 1.5 Flash outperforms GPT and Claude models on MRCR and IDK at long enough context lengths, and that advantage is retained up to 1M tokens of context.

This behavior highlights a potential tradeoff between short-context performance and long-context performance that these models may make. On MRCR, we note that GPT and Claude models outperform Gemini models on context smaller than 8K, but have a correspondingly higher rate of decay in length generalization performance, indicating that short and long context performance may be difficult to balance.

\subsection{Parallel Model Family Curves on MRCR}

In Figure~\ref{fig:fig2}, we pointed out that various model families have parallel curves on MRCR, an intriguing phenomenon. We highlight this property as both evidence for the fact that the evaluation bears high signal, as well as a pointer to paths for future investigation. Since the curves between certain model pairs so closely parallel each other, we suspect that there were uniquely similar aspects of the model training process in these models (even though there may be an absolute disparity in performance). Future work should investigate evaluations which reveal implicit information about model training.

We also note that we found MRCR to be a particularly robust measure of performance. MRCR behavior across models required no changes to the prompt in order to ensure the model gave a verifiable output with low variance.

\subsection{Stratifying Latent List by Complexity}
\label{subsec:complexity_stratification}

\begin{figure}[h]
    \centering
    \includegraphics[width=0.8\textwidth]{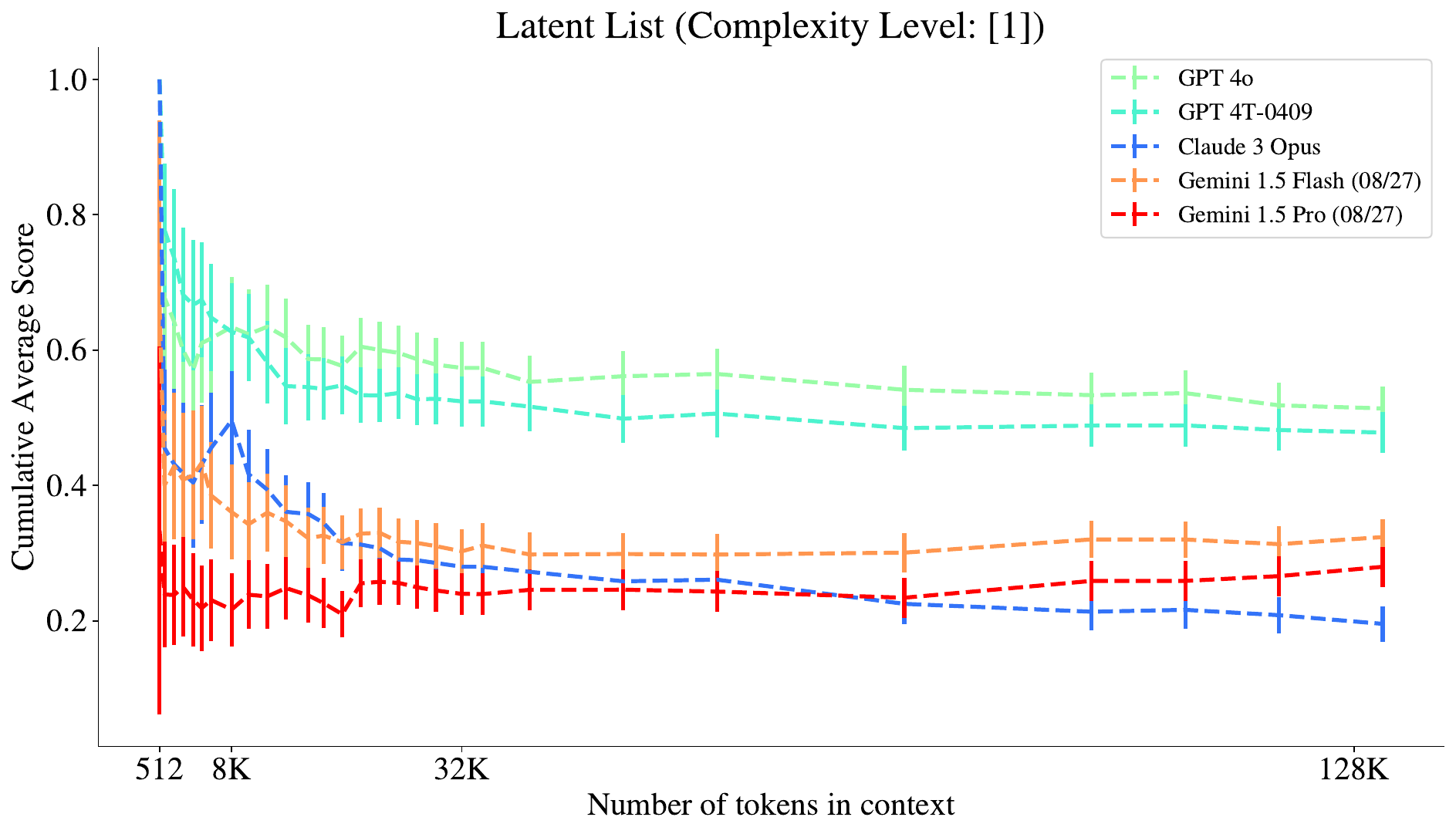}
    \caption{Latent List: Complexity Level $1$}
    \label{fig:fig4a}
\end{figure}

\begin{figure}[h]
    \centering
    \includegraphics[width=0.8\textwidth]{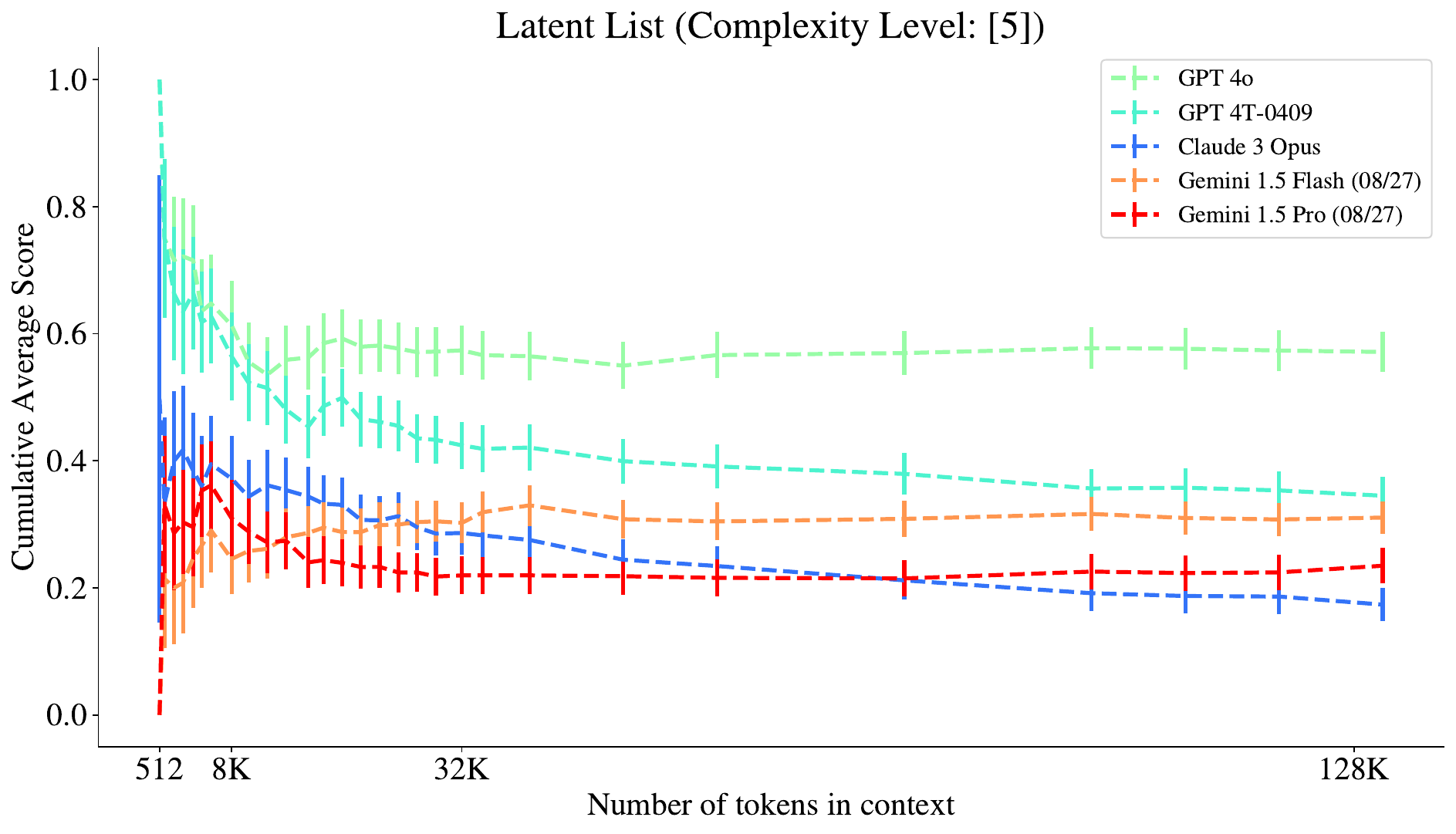}
    \caption{Latent List: Complexity Level $5$. At 128K, Gemini and GPT-4T have similar performance.}
    \label{fig:fig4b}
\end{figure}

\begin{figure}[h]
    \centering
    \includegraphics[width=0.8\textwidth]{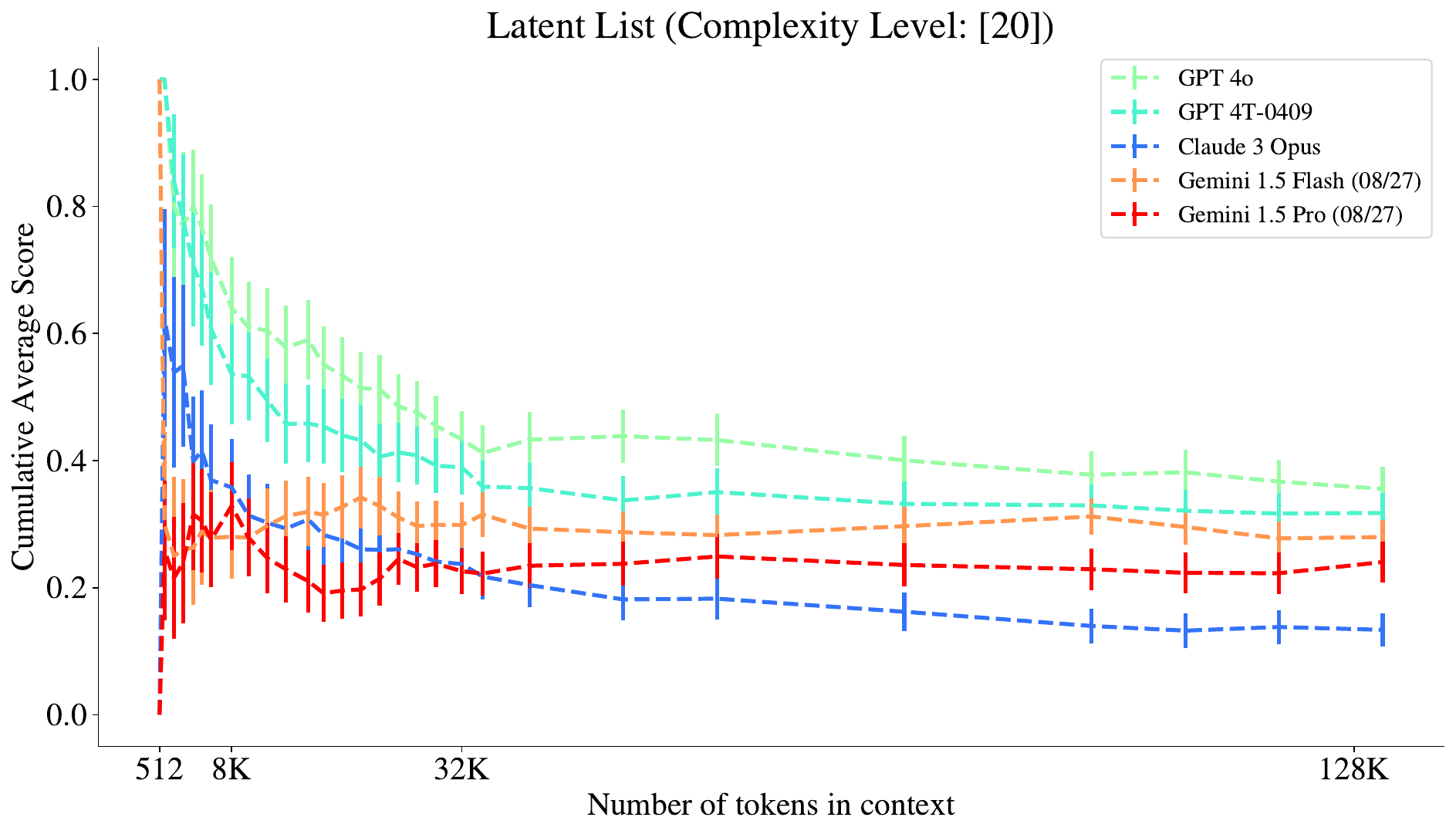}
    \caption{Latent List: Complexity Level $20$. At 128K, Gemini and both GPT models have similar performance.}
    \label{fig:fig4c}
\end{figure}

In this section we validate the claim that higher complexity tasks are more difficult as measured by score (Figures~\ref{fig:fig4a}, \ref{fig:fig4b}, \ref{fig:fig4c}). We also note that the highest complexity task has a significantly steeper slope as a function of context length, especially for the highest performing model, GPT-4o, suggesting that as the task complexity increases, the more rapid the performance fall-off as a function of context length. Notably, on the highest task complexity, performance of GPT-4o is on par with GPT-4T and the Gemini 1.5 models (Figure~\ref{fig:fig4c}).

\subsection{Slicing IDK Performance When the Answer is ``I Don't Know''}

\begin{figure}[h]
    \centering
    \includegraphics[width=0.8\textwidth]{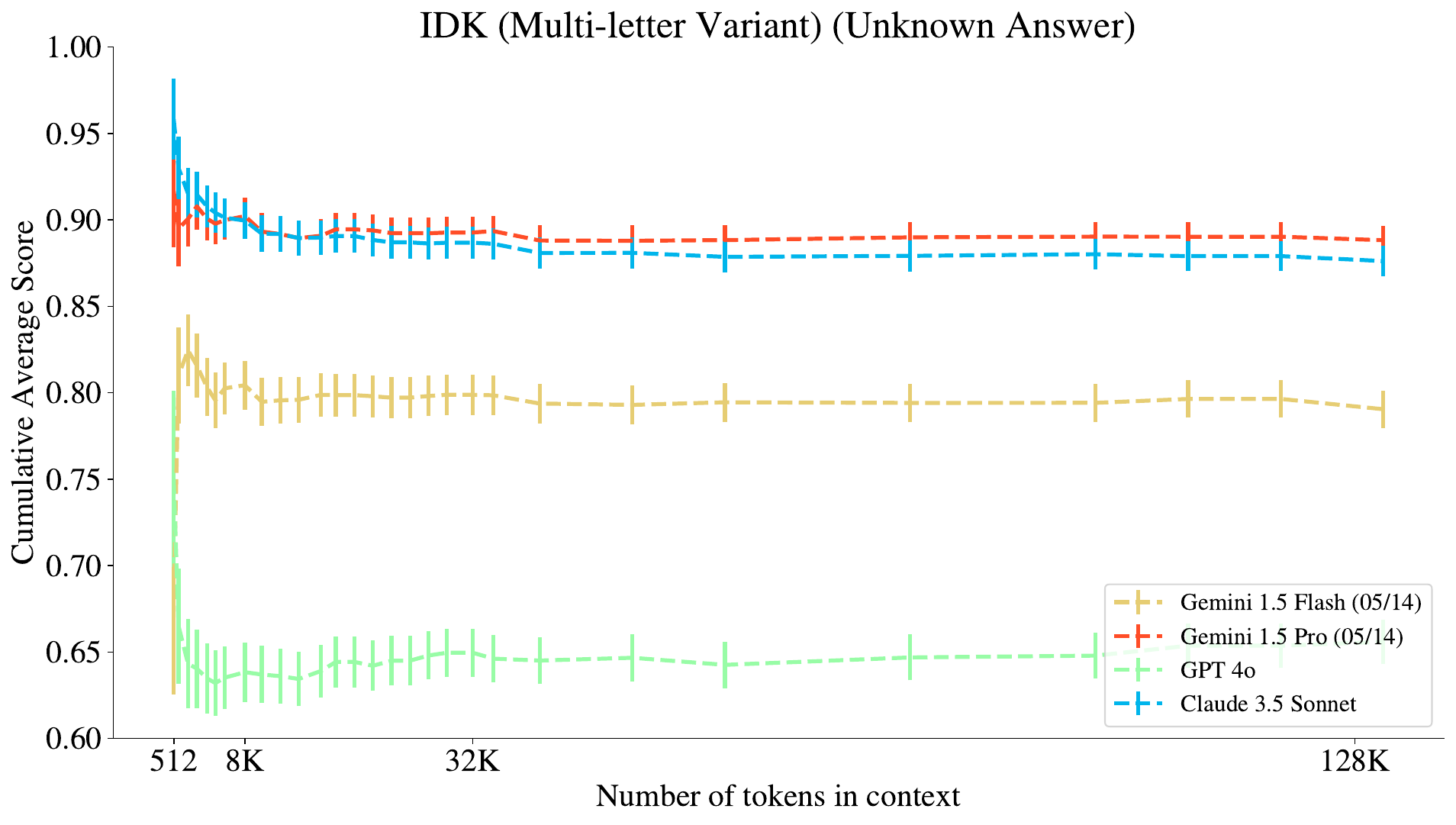}
    \caption{IDK: The Answer is Always Unknown. At around 8K context, Gemini 1.5 Pro overtakes Claude 3.5 Sonnet as the top performer on this subtask.}
    \label{fig:fig4d}
\end{figure}

In Figure~\ref{fig:fig4d} we similarly observe a steeper fall-off curve in performance when we restrict IDK to task instances where the answer is only ``I don't know''. This task is harder because it requires the model to assess the full context for information which may not be present, rather than identify a single location where an answer may be located. Notably, Gemini 1.5 Pro performs slightly better than Claude 3.5 Sonnet on long-enough contexts on this subtask.

\subsection{An Interesting Failure Mode: GPT Models Hallucinate Puzzles in IDK}

We previously noted that the GPT-4 family performs significantly worse than the other model families on IDK in Figure~\ref{fig:fig4d} – this is due to an interesting quirk in that model family’s behavior. When presented with a string of random letters, both GPT-4 models sometimes assume that there is a hidden riddle in the text, and attempt to ``solve" the riddle by hallucinating the presence of one of the answer choices. This leads to a significantly worse performance. We provide examples of this behavior in Figure~\ref{fig:gpt4o_idk}.

Note that the OpenAI API does not allow inputs with simpler context (like for instance, ``X X X$\cdots$''), presumably due to repetition attack schemes \citep{nasr2023scalable} which cause the model to reproduce its training data.

\begin{figure}[h]
\centering
\begin{tcolorbox}[
  enhanced jigsaw, % For the puzzle piece effect
  colback=gray!10, % Light background
  colframe=gray!50, % Medium border
  sharp corners,
  boxrule=1pt, % Thicker border
  arc=3mm, % Rounded corners
  fontupper=\small\ttfamily, % Typewriter font for everything
  toptitle=1mm,
  bottomtitle=1mm,
  title=Failure Modes for GPT-4o on IDK
]

\begin{tcolorbox}[
  colback=cyan!10, % Cyan background for prompt title
  colframe=cyan!50, % Cyan border for prompt title
  width=0.95\linewidth, % Make it slightly narrower
  title=IDK Example 1,
  left=3mm, right=3mm % Add small margins on the sides
]
A man is walking down the street when he sees a friend. The friend is wearing a Stetson hat and Oakley sunglasses. W F D O F J F J U Q C M Z J U A G O C E W W H R S L E E U W I L  $\cdots$

$\vdots$

Question: What is the friends name?

Choices:

(A) John Oakley

(B) Jim Stetson

(C) Bob Field

\colorboxed{red}{\texttt{(D) I don't know}}

Answer:
\end{tcolorbox}

\begin{tcolorbox}[
  colback=green!10, % Green background for context
  colframe=green!50, % Green border for context
  width=0.95\linewidth,
  title=GPT-4o's Incorrect Response
]
The friend's name can be deduced from the description given. The friend is wearing a Stetson hat and Oakley sunglasses. The choices provided are: (A) John Oakley (B) Jim Stetson (C) Bob Field (D) I don't know Given the clues: - Stetson hat (suggesting the name Stetson) - Oakley sunglasses (suggesting the name Oakley) The correct answer is likely to be a combination of these clues. Therefore, the friend's name is: (B) Jim Stetson 
\end{tcolorbox}

\begin{tcolorbox}[
  colback=cyan!10, % Cyan background for prompt title
  colframe=cyan!50, % Cyan border for prompt title
  width=0.95\linewidth, % Make it slightly narrower
  title=IDK Example 2,
  left=3mm, right=3mm % Add small margins on the sides
]
I am searching for jobs in the computer technology industry. X F R L L O G W J Z D V F X E G Z W G E A L C R W N T E D A F Y $\cdots$

$\vdots$

Question: What company did I not apply too?

Choices:

(A) Google 

(B) Meta 

(C) Apple

\colorboxed{red}{\texttt{(D) I don't know}}

Answer:
\end{tcolorbox}

\begin{tcolorbox}[
  colback=green!10, % Green background for context
  colframe=green!50, % Green border for context
  width=0.95\linewidth,
  title=GPT-4o's Incorrect Response
]
Based on the provided string, which appears to be a word search puzzle, the task is to identify which of the companies listed in the choices (Google, Meta, Apple) does not appear in the puzzle. Let's look for each company name in the string: 1. **Google**: - Searching for "GOOGLE" in the string, we find "GOOGLE" at the position starting from the 67th character: `...G O O G L E...` 2. **Meta**: - Searching for "META" in the string, we find "META" at the position starting from the 626th character: `...M E T A...` 3. **Apple**: - Searching for "APPLE" in the string, we do not find the sequence "APPLE" anywhere in the string. Given this information, the answer is: (C) Apple
\end{tcolorbox}

\end{tcolorbox}
\caption{Examples for which GPT-4o fails on IDK tasks due to hallucinating puzzles.}
\label{fig:gpt4o_idk}
\end{figure}

\section{Related Work}
\label{sec:related}

\subsection{Broader Work on Long Context Evaluation}

For the purposes of this paper, we consider ``long-context'' to mean at least 32K context examples (ideally at least 128K context). There have been many recent long-context evaluation benchmarks, which tend to focus on retrieval \citep{needle, geminiteam2024gemini15unlockingmultimodal, li2024needlebenchllmsretrievalreasoning, hsieh2024rulerwhatsrealcontext, zhang2024infty, kovcisky2018narrativeqa, bohnet2024longspanquestionansweringautomaticquestion} and in-context many-shot learning \citep{agarwal2024manyshotincontextlearning, bohnet2024exploringbenchmarkingplanningcapabilities} rather than reasoning. There is also existing work on long-context summarization \citep{chang2024booookscore, kim2024fablesevaluatingfaithfulnesscontent}. In this paper we focus exclusively on evaluations for reasoning over a long-context. In particular, we assume that there is a sparse set of relevant information embedded in the context, not all of which is immediately identifiable independent of other portions of the context. The relevant information must be synthesized rather than merely retrieved.

Another popular approach for claiming long-context capability is to create plots of context length vs. average perplexity on a designated set of long documents \citep{claude, geminiteam2024gemini15unlockingmultimodal}. It is worth noting that the decreasing perplexity for a fixed model (the difference in magnitude between the average perplexities at the longest context and the shortest context) is anti-correlated with the difference in error between the longest context and the shortest context on the long reasoning evaluations we introduce in this report. Note that the anti-correlation is immediate from the fact that the perplexity plots are approximately monotone decreasing as a function of context length, while the error plots for the evaluations we present are approximately monotone increasing as a function of context lengths (accuracy is approximately monotone decreasing). Thus, examining the context-vs-perplexity plot on a single model may not be a proxy for understanding model performance as measured by accuracy on complex long reasoning tasks, though it is possible that it can be a proxy for capabilities like many-shot in-context learning for translation tasks (see for instance some of the examples in \cite{geminiteam2024gemini15unlockingmultimodal}).

Another noteworthy point is that the majority of long-context evaluations use multiple-choice formats (sometimes only with two possible responses) -- this paradigm is considerably limiting. While some other methods use ROUGE-style scores (especially in summarization) or human evaluations, both of these approaches can be additionally problematic in both noisiness (see for instance the discussions in \cite{cohan-goharian-2016-revisiting} and \cite{akter-etal-2022-revisiting}) and cost.

\subsection{Challenges with Developing Reliable and Minimal Long Reasoning Evaluations}

There are a few pre-existing evaluations which study arbitrarily long reasoning tasks, with evaluations sometimes going up to 1M context, including \cite{zhang2024infty, hsieh2024rulerwhatsrealcontext, li2024needlebenchllmsretrievalreasoning,kuratov2024babilongtestinglimitsllms, loft}. These existing evaluations tend to suffer from at least one of a few deficiencies:
\begin{itemize}
    \item \textbf{Short-circuiting:} When an evaluation task does not account for information in either training data or later in the context which the model can use to avoid requiring the full context to answer a query, we say the model can ``short-circuit'' the long-context and perform well on the task while not actually representing a utilization of the context. As one example, the tasks in \cite{kuratov2024babilongtestinglimitsllms} may all suffer from this problem, both because the evaluation from which the expanded tasks are derived is heavily leaked and because the tasks themselves are significantly flawed \citep{kaushik-lipton-2018-much}. Many similarly constructed evaluations (perhaps using different base evaluations) have the same issues.
    \item \textbf{Secret retrieval tasks:} Some recently popular evaluation benchmarks (for instance, \cite{li2024needlebenchllmsretrievalreasoning, hsieh2024rulerwhatsrealcontext}) describe a subset of their tasks as testing long-context reasoning capabilities. However, due to their construction, they often do not represent a task which requires anything more than single-needle or multi-needle retrieval. As one example of such a task, we consider the Variable Tracing (VT) task from RULER \citep{hsieh2024rulerwhatsrealcontext}. On the surface level, the task contains a context filled with many variable assignments (e.g. $X = 3; Y = X; Z = Y$ and so on, interspersed with distractor context). The model is then required to enumerate all variables which have the value $3$. However, in the default RULER implementation of this task, every variable which has been introduced in the context actually indeed has the value $3$ - there are no spurious variables present. Thus this setting of the task is ultimately reduced to a multi-needle retrieval task, where the needles correspond to the mentioned variable names, and is why we see exceedingly high performance on this task from many models.
    \item \textbf{Out-of-distribution distractor context:} Many existing tasks proceed by inserting dramatically out-of-distribution context in order to form the irrelevant information component of the evaluation, such as Paul Graham essays or repeated phrases \citep{hsieh2024rulerwhatsrealcontext, li2024needlebenchllmsretrievalreasoning, needle}. This setup makes the problem significantly easier, since it implicitly brings any long reasoning task closer to a retrieval task -- if the relevant information is \textit{a priori} identifiable without understanding anything about the interrelated nature of the relevant information, then the task becomes effectively a multi-needle retrieval task.
    \item \textbf{Training on toy synthetic tasks:} Some evaluation methods (often older evaluation sets from the pre-modern LLM era, like the Long Range Arena (LRA) \citep{tay2020long}) require the model to train on the task to see good performance on the task. These evaluations test the ability of a given architecture to \textit{learn} a specific task, which while potentially interesting, is not the key point of interest in our setting, as we hope that via next-token prediction we will learn far more interesting long reasoning circuits that apply more generally over language and multimodal elements.
\end{itemize}
See Appendix~\ref{app:related} for a more detailed breakdown of these issues and how they apply to additional existing proposed long reasoning evaluations.

Another relevant long-reasoning evaluation is the SPIDER task in the LOFT benchmark (\cite{loft}), which tests multi-hop reasoning for SQL queries. SPIDER is quite focused on a particular application (SQL), and we include some orthogonal tests of long reasoning in our benchmark.

\evalname avoids the above pitfalls by ensuring that the irrelevant information in the context is indeed completely unrelated to the information required to answer the query, and by ensuring that the relevant information in the query is completely unique and/or there is no information present in training data which would be helpful for answering the query, since it is too generic. For most of the tasks, we also ensure that the irrelevant information is not significantly out-of-distribution compared to the relevant information, thus creating a more realistic setting where the relevant information is not clearly different from surrounding context.

\section{Conclusion}
\label{sec:conclusion}

We introduced the \evalname evaluations, a suite of long-context synthesis and reasoning evaluations built using the \frameworkname framework for developing long-context reasoning evaluations which can be arbitrarily extended in length and set to arbitrary complexity levels while avoiding leaked context from pre-existing evaluations. We propose three minimal simple tasks in this framework -- Latent List, MRCR, and IDK -- which highlight simple cases of tasks beyond retrieval which current frontier long-context models do not solve, even up to 32K context. We analyze ten frontier models with long-context capabilities on these tasks up to 128K context, and we further run these evaluations on the Gemini 1.5 family up to 1M context, and observe that though there is a significant initial degradation in behavior on these evaluations, after a certain point, many (but not all) frontier models experience a non-trivial flattening of the context-vs-performance curve, suggesting that while some long-context capabilities are present in these models (e.g. the ability to retrieve), there is still a significant gap in capability compared to the goal of reasoning over long contexts.

One important part of the contributions of this work is to validate the robustness and smoothness of MRCR across multiple frontier models. MRCR benefits from using a very simple metric and is more robust to prompting choices as compared to Needle-in-a-Haystack. In internal testing, we have also discovered that this evaluation has high signal for smaller models as well as models which have not been post-trained. Thus, we also make a recommendation -- given that MRCR minimally tests a simple primitive of beyond-retrieval understanding of long contexts while also capturing harder distractor contexts for retrieval, MRCR and its natural extensions are a suitable default replacement for the popular Needle-in-a-Haystack evaluation, in which the needle is often quite apparently different from the rest of the text in a manner that stands out and which fails to capture nuance in measuring model capabilities over long-contexts.

\section{Contributions}
\label{sec:contributions}

\textbf{Corresponding Author}: Kiran Vodrahalli\\
\textbf{Core Contributors}: Santiago Onta\~{n}\'{o}n, Nilesh Tripuraneni, Kelvin Xu, Sanil Jain, Rakesh Shivanna, Jeffrey Hui, Nishanth Dikkala, Mehran Kazemi, Bahare Fatemi\\
\textbf{Contributors}: Rohan Anil, Ethan Dyer, Siamak Shakeri, Roopali Vij, Harsh Mehta, Vinay Ramasesh, Quoc Le, Ed Chi, Yifeng Lu, Orhan Firat, Angeliki Lazaridou, Jean-Baptiste Lespiau, Nithya Attaluri, Kate Olszewska

\noindent The roles are defined as below:
\begin{itemize}
    \item Corresponding Author: Led the project, created many of the evaluations and the overall framework.
    \item Core Contributor: Individual who had significant impact throughout the project.
    \item Contributor: Individual that had partial contributions to the project and was partially involved with the effort.
\end{itemize}

\noindent Within each category, the authors are presented in no particular order.

\bibliography{main}

\appendix

\section{Appendix: Prompts}
\label{app:prompts}

\subsection{Latent List}

The prompt format for Latent List is:
\begin{lstlisting}
Pretend to be a Python interpreter. You will see a sequence of updates which correspond to list operations. Here are some examples.
===================
 Example 1:
 
  >> a = [1, 2, 3, 4, 5, 6]
  >> a.pop(0)
  >> a.pop(4)
  >> a.remove(3)
  >> a.sort()
  >> a.sort()
  >> a.append(1729)
  >> a.sort()
  >> a.append(1273)
  >> {view_op}(a[1:3])
  Output: {ans1}

===================
 Example 2:
 
  >> a = [1, 2, 3, 4, 5, 6]
  >> a.insert(3, 3129)
  >> a.pop(2)
  >> a.append(-4610)
  >> a.remove(2)
  >> a.sort()
  >> a.reverse()
  >> a.reverse()
  >> a.sort()
  >> {view_op}(a[1:2])
  Output: {ans2}

===================
 Example 3:

  >> a = [1, 2, 3, 4, 5, 6]
  >> {inputs}
  >> ...
  >> {view_op}(a[{list_slice}]
  Output:
\end{lstlisting}

\subsection{MRCR}

The prompt format for MRCR is:
\begin{lstlisting}
Here are some examples of conversations succeeded by a follow-up question answered correctly:
======== EXAMPLE 1 ========
User: {user_query11}
Model: {model_output11}
User: {user_query12}
Model: {model_output12}
User: Add the sentence {random_string} to the {key1}.
Model: {random_string} {correct_model_output1}
======== EXAMPLE 2 ========
User: {user_query21}
Model: {model_output21}
User: {user_query22}
Model: {model_output22}
User: Add the sentence {random_string} to the {key2}.
Model: {random_string} {correct_model_output2}
======== EXAMPLE 3 ========
User: ...
Model: ...
...
User: {user_query31}
Model: {model_output31}
...
User: ...
Model: ...
...
User: {user_query32}
Model: {model_output32}
...
User: ...
Model: ...
User: Add the sentence {random_string} to the {key3}.
Model:
\end{lstlisting}

\subsection{IDK}

An example prompt for IDK is

\begin{lstlisting}
A group of people are gathered around a table at a park.  T P L J B O I Q Y N V Z I R E L X Q O G N U C J O L E L U A Y E Y R R K Z G A E I Q E R L Q V D N F I O X E Y Y X R O X P H S G O Z P X Q T Z U J C B K V V W T E K J F I N S H Y O O A D Y G N T U Y L Y M C O H V P U X Z M F A P I K T C J A K M C Z V X I H D J B M I G U N P C K F E D J Y N K X B F G B Q H C J N I P L M N I P C B W Y F E P F C F M C A L A M H X X P X W B U O S Z X S M M S N A C U Y N C Q U B F W V P R N L R N F C V D G C H J E G Z Q P I Q W S C B Y Q G E U H W H O C H L C A B N Y W B K F K B X K K G P F H P D O O N A D O Y O T A X I R S B Z A B D T B M G C S O E S V K D J F D I N Y C D N Y Q Z O V O B H S Q S W G Y R B J U C B B J Q E H G Q F O Z Z T N Q O F G E Y P U D W M S D O L N Q A B D B H W R M C P K Z Y V M G L S I B I U F X O M Y F I C Z H I K G I A B N B A Q L A O C O R W D Q U R J V Z V T L V Q G H S J M P Z H E W H T S X X B F D X F H G V P X X A L T T O W X O J Y W K U R P M V S O B F O J E P A O W I Y U Y K W M X R K J C N O C M F C Q C K Q K C E S E F B A.
Question: What city is the park located in?
Choices:
(A) Fresno
(B) Berkeley
(C) Sacramento
(D) I don't know
Answer:
\end{lstlisting}

\section{Appendix: Chance Rate Estimation}

\subsection{Estimating the Chance Rate for Latent List}

Since Latent List is scored via an approximate metric, it is non-immediate to determine what a reasonable estimate of a chance rate for this evaluation is. To estimate the chance rate, for each view operation type and for each difficulty level, we assume the random model uniformly samples from the space of all possible outputs and compute the average score. In particular, for the \texttt{len} operation, we suppose the random model uniformly outputs a length between $0$ and the number of relevant operations (corresponding to difficulty level), and that the true length is similarly uniformly distributed over this range.  With no further assumptions about the random model, the chance rate is $0.01\%$. Therefore, to provide a more realistic estimation of the chance rate, for the \texttt{print}, \texttt{sum}, \texttt{max}, \texttt{min} operations, we assume the random model instead may restrict to considering the specific numbers present in the relevant operations, though it may not necessarily know how to combine them to compute the correct latent list. In particular, for \texttt{print} and \texttt{sum}, we assume the random model uniformly samples a subset of the numbers involved in the relevant operations, and also subsamples from the initial list (the array \texttt{[1, 2, 3, 4, 5, 6]}) to construct the random model's guess for the value of the latent list (randomly shuffled). Then, for \texttt{print} and \texttt{sum}, the random model samples a slice of that list to compare to the true slice, and also computes the sum over that slice to compare to the true sum. For \texttt{min} and \texttt{max}, we assume the random model uniformly picks a single entry of the randomly selected latent list and compares to the true list. We then average over the complexity levels (complexities $1, 5, 20$), which are uniformly present in the evaluation instances. Computing the chance rate in this manner yields a chance rate of $12.2\%$ on average over all complexity levels, $16.9\%$ for complexity level $1$, $11.3\%$ for complexity level $5$, and $8.5\%$ for complexity level $20$.

\subsubsection{Estimating the Chance Rate for MRCR}

\begin{figure}
    \centering
    \includegraphics[width=0.8\textwidth]{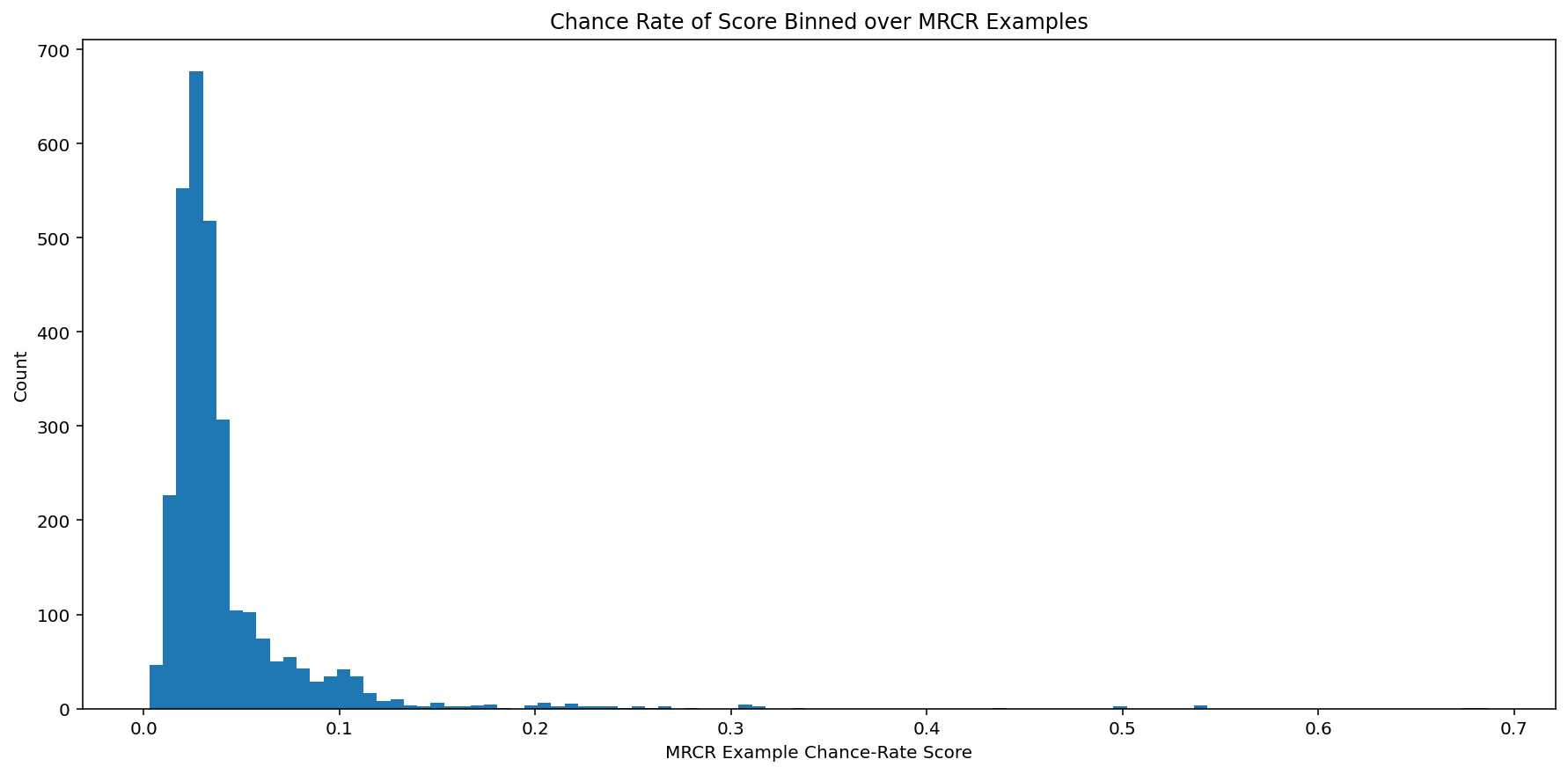}
    \caption{We plot the histogram of chance rates per example for MRCR as computed by the simple random model which uniformly outputs one of the pieces of writing, as computed with the approximate edit distance metric.}
    \label{fig:mrcr_chance_rate_basic}
\end{figure}

Since this evaluation is scored via an approximate edit distance metric, it is non-immediate to determine what a reasonable estimate of a chance rate for this evaluation is. One approach we can take to determine an estimate of the chance rate is to assume that the model outputs one of the most related options (however, this does not take into account the cases where the model may partially reconstruct the correct output, where the model may refuse to reproduce any output, and where the model outputs a less relevant output). The chance rate when computed assuming a model randomly outputs any of the possible outputs in the conversation history is $4\%$, and the histogram of chance rates per example is plotted in Figure~\ref{fig:mrcr_chance_rate_basic}.

\begin{figure}
    \centering
    \includegraphics[width=0.8\textwidth]{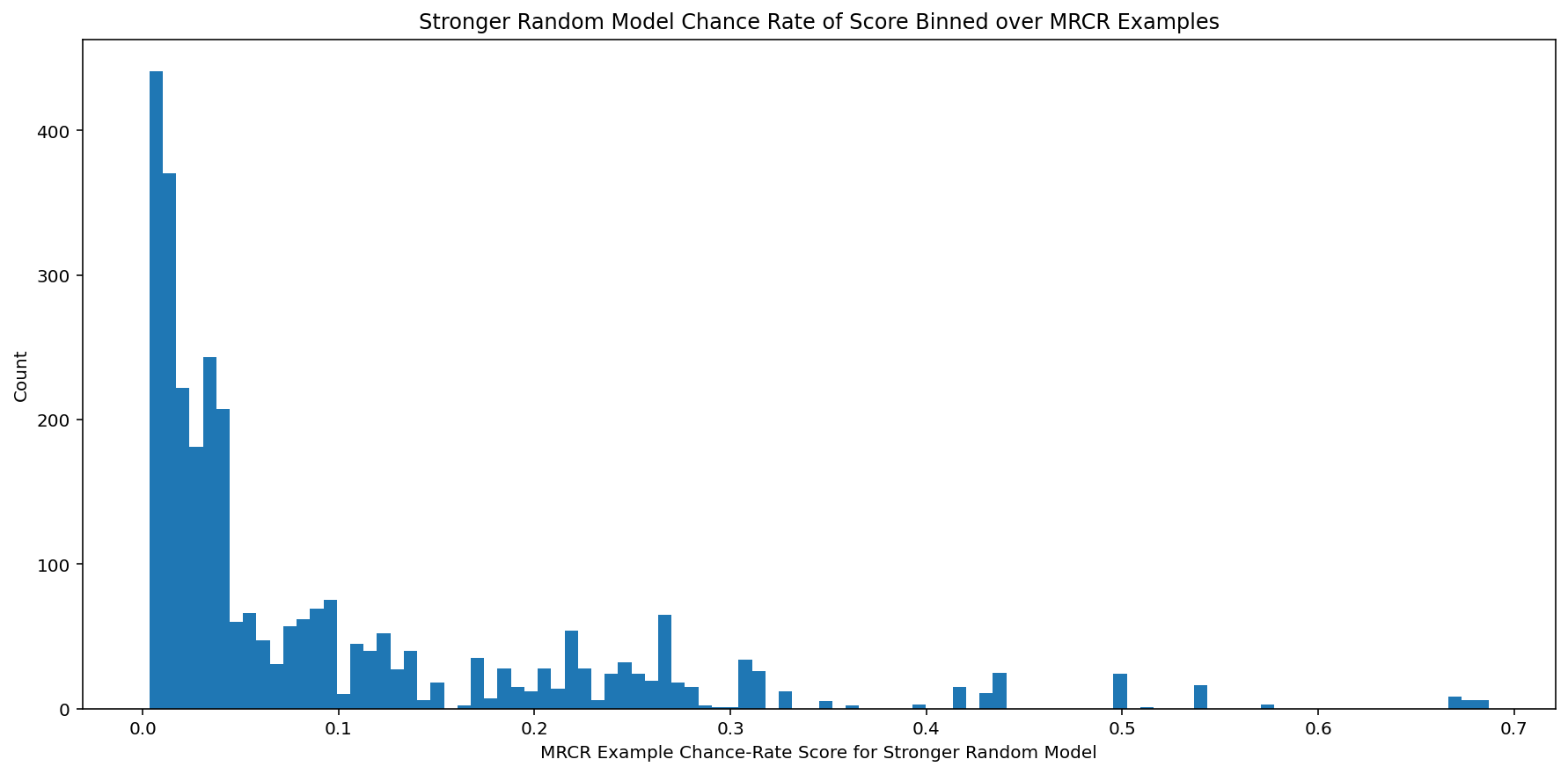}
    \caption{We plot the histogram of chance rates per example for MRCR as computed by the stronger random model which uniformly outputs one of the pieces of writing which has at least one of the topic or format matched to the desired answer, as computed with the approximate edit distance metric.}
    \label{fig:mrcr_chance_rate_stronger}
\end{figure}

We can also consider computing the chance rate over instances where we assume a random model would output one of the instances where at least one of either the topic or the format is matched. The chance rate when computed this way is $9\%$, which as expected is larger. The histogram of chance rates per example is plotted in Figure~\ref{fig:mrcr_chance_rate_stronger}.

\subsection{Chance Rate for IDK}

Since the query is a multiple choice question with four options, the chance rate is $25\%$.

\section{Appendix: More Plots}
\label{app:more_plots}

\begin{figure}
    \centering
    \includegraphics[width=0.8\textwidth]{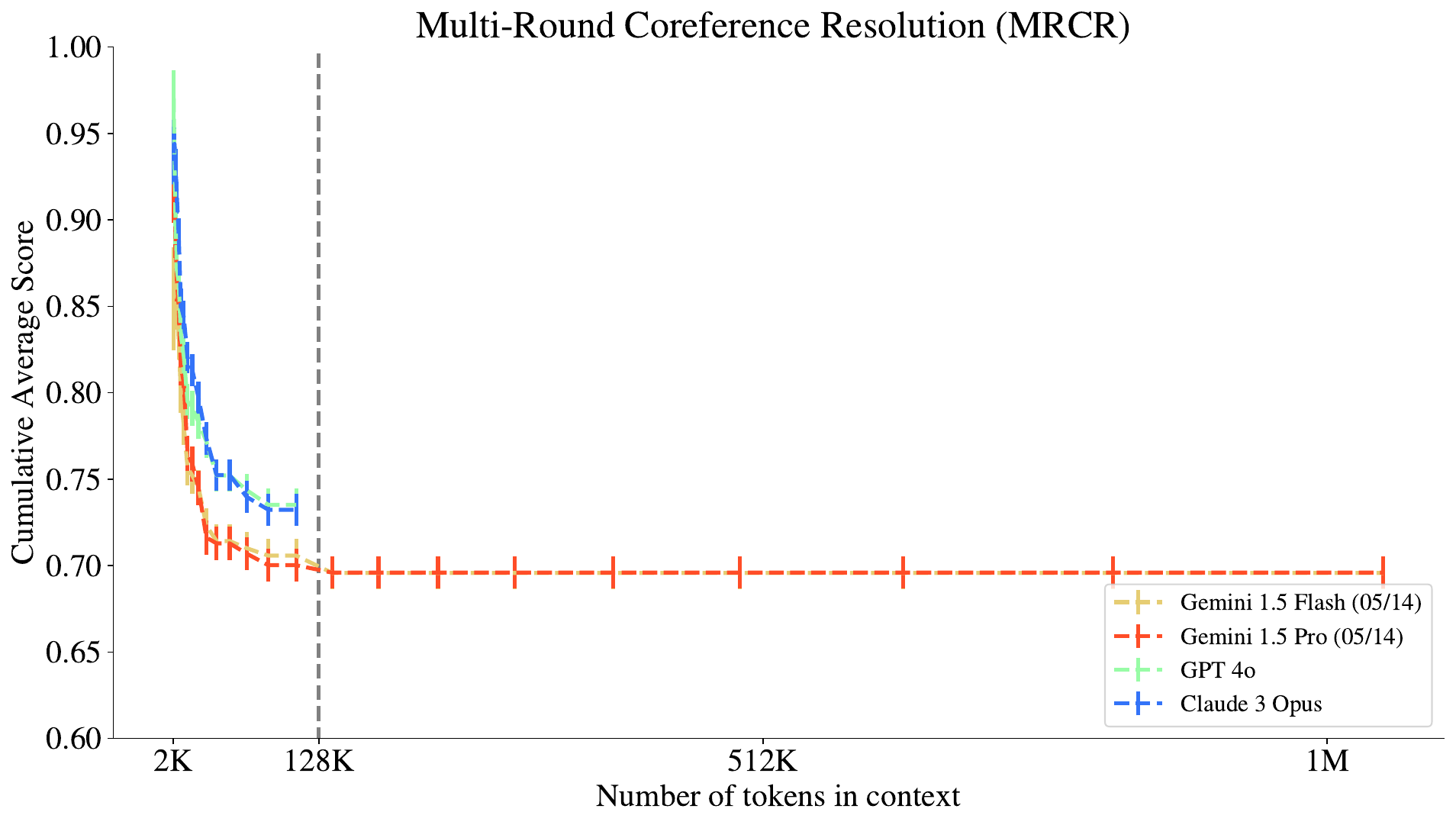}
    \caption{MRCR Performance with older Gemini models. Here we observe that more recent Gemini models represent a significant improvement in performance on this task. In Figure~\ref{fig:fig1}, Gemini 1.5 Pro (08/27) now outstrips all competitor models after around 8K context, and Gemini 1.5 Flash (08/27) outstrips all competitor models after around 20K context.}
    \label{fig:fig5p1b}
\end{figure}

\begin{figure}
    \centering
    \includegraphics[width=0.8\textwidth]{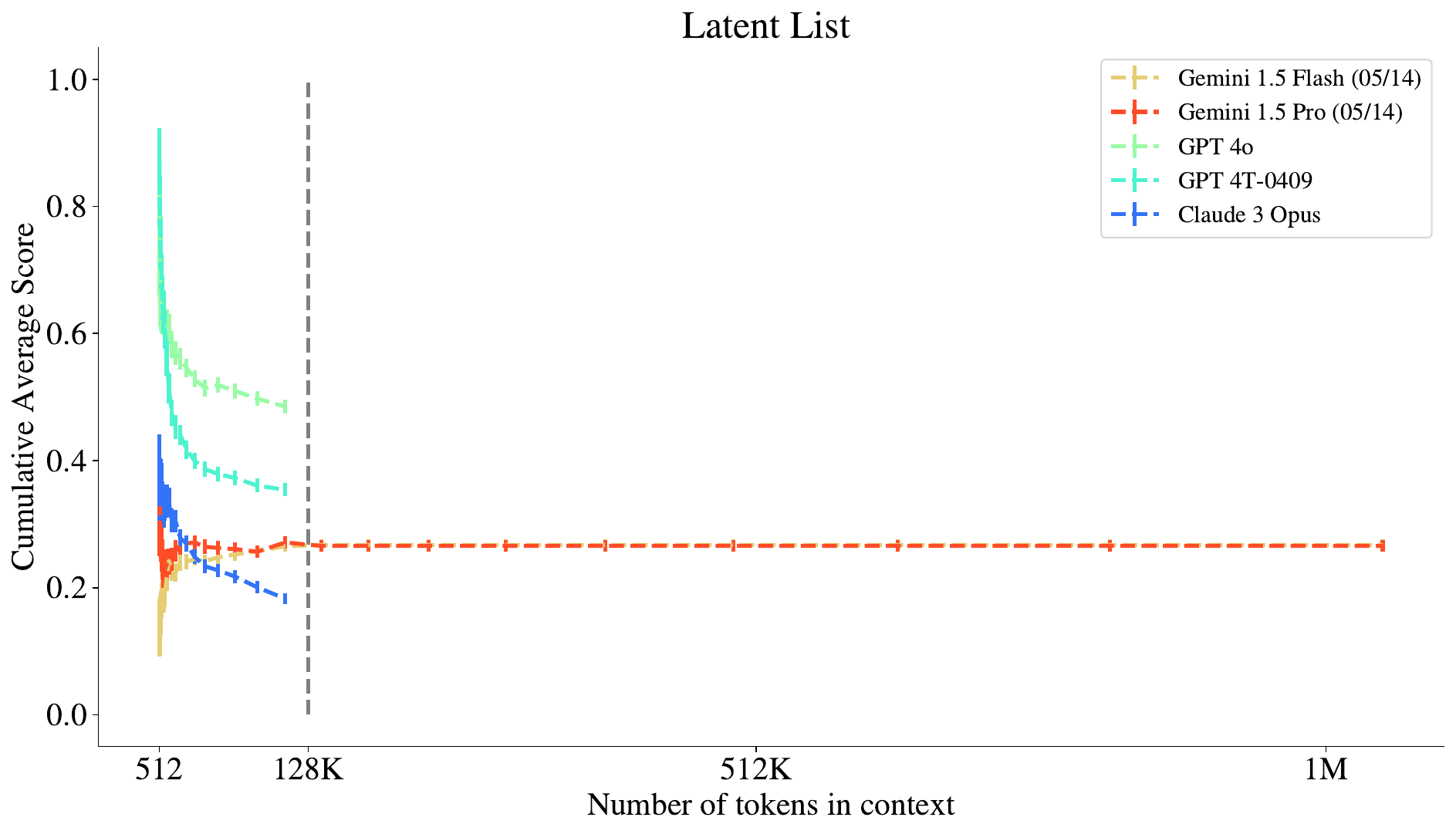}
    \caption{Latent List Performance with older Gemini models. Here we observe that the more recent Gemini models represent a significant improvement on this task, particularly for Gemini 1.5 Flash, which has poor performance in the short-context setting on this task. The performance generalization trend up to 1M context is unaffected.}
    \label{fig:fig5p2b}
\end{figure}

\begin{figure}
    \centering
    \includegraphics[width=0.8\textwidth]{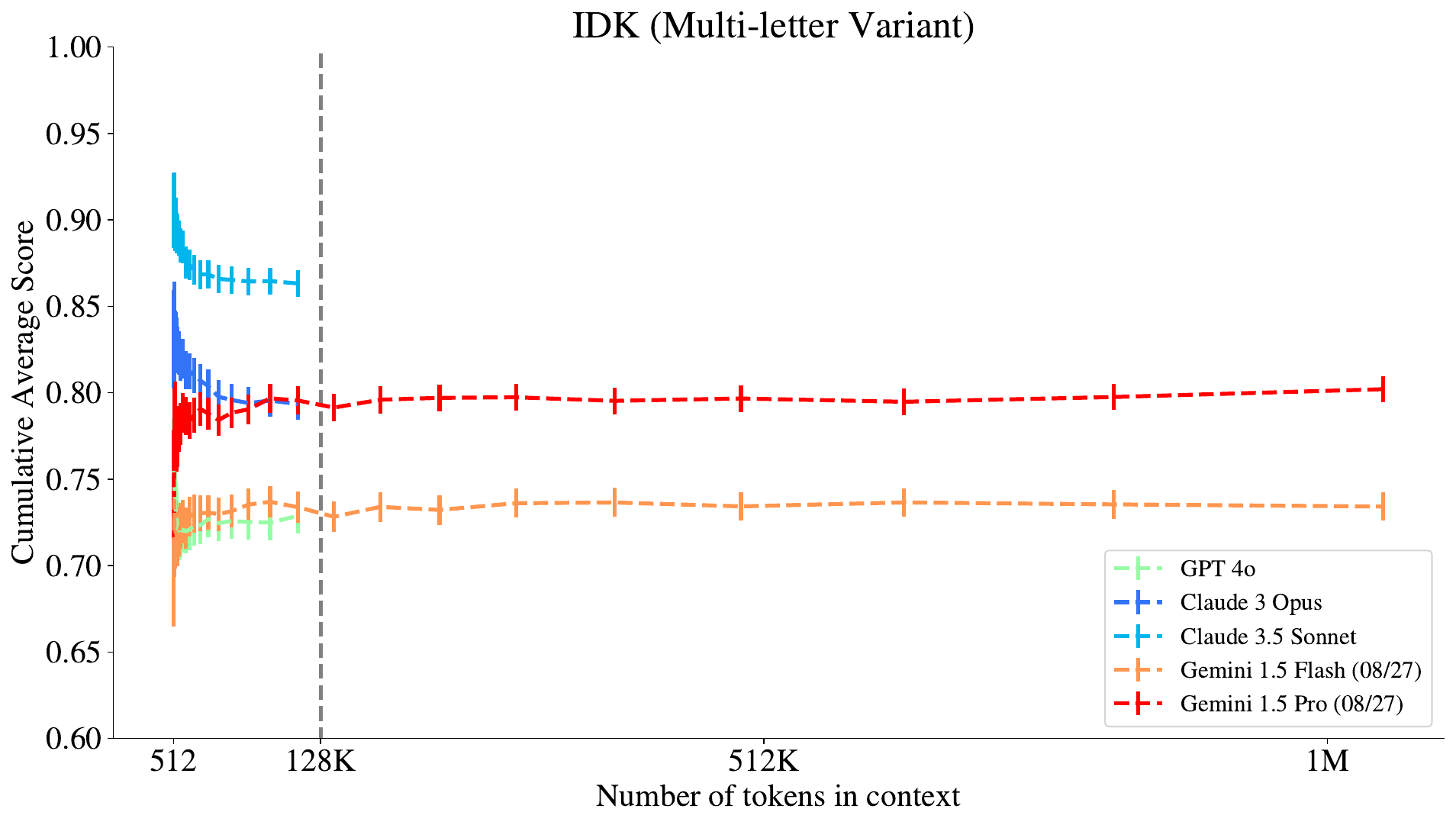}
    \caption{IDK Performance with more recent Gemini models. Here we observe some regressions on IDK for more recent Gemini models, but the general trend of performance generalization across context lengths up to 1M context remains the same.}
    \label{fig:fig5p3b}
\end{figure}

\begin{figure}
    \centering
    \includegraphics[width=0.8\textwidth]{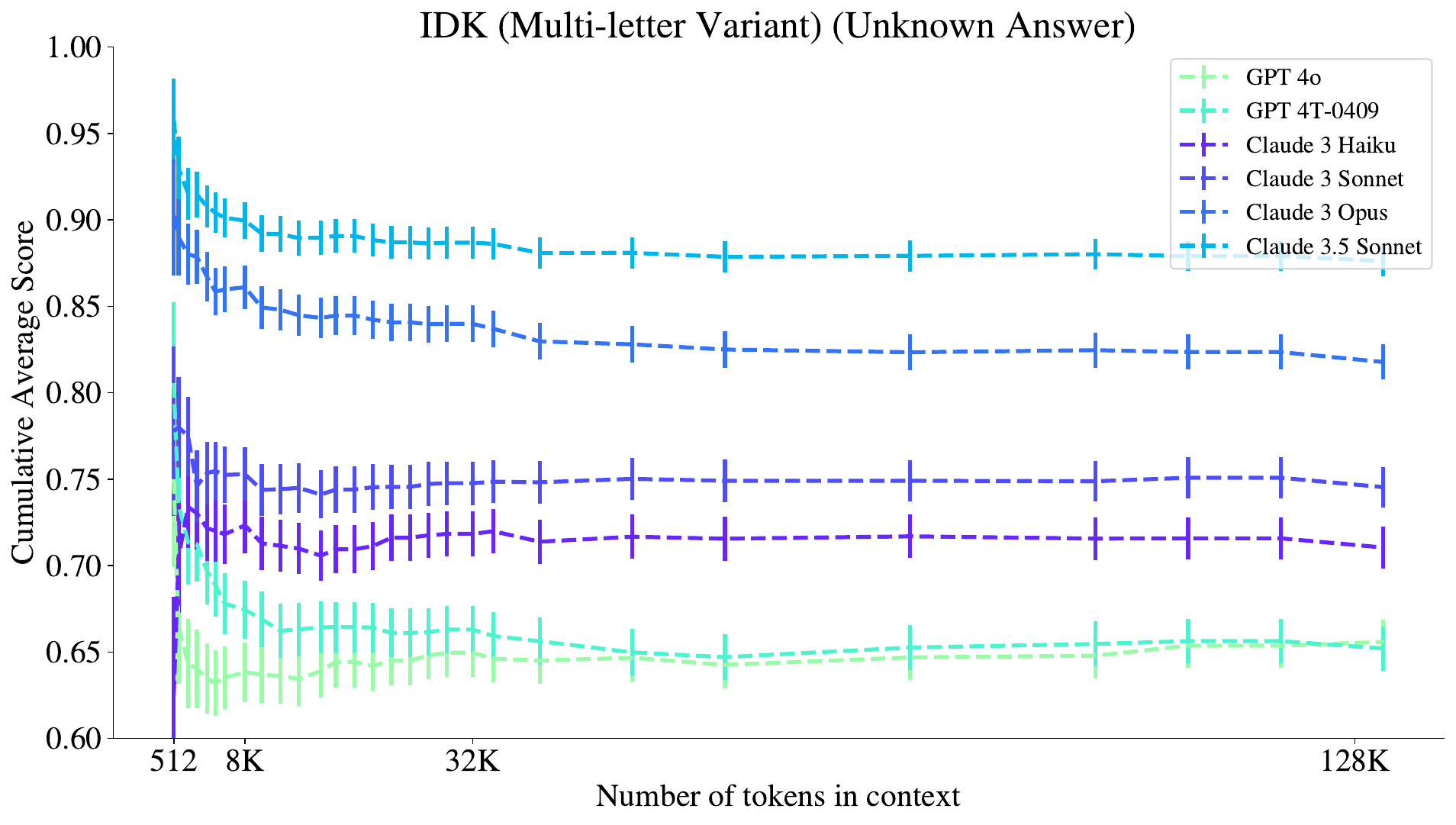}
    \caption{IDK Performance with all GPT and Claude models, restricted to the ``Unknown Answer'' setting, where the answer is always ``I don't know.'' We observe that both GPT models perform poorly on this task, and GPT-4o is a regression compared to GPT-4T on this task, indicating it is more certain of an answer when there is none. For Claude models, we observe that the performance trends up as a function of model size, with the exception of Claude-3.5 Sonnet, which significantly outperforms the second best Claude model (Claude-3 Opus).}
    \label{fig:fig5p4c}
\end{figure}

In this section, we include some additional plots (Figures~\ref{fig:fig5p1b}, \ref{fig:fig5p2b}, \ref{fig:fig5p3b}, \ref{fig:fig5p4c}) for the Michelangelo tasks on different sets of models which do not necessarily have the best performance. With this section, we have full coverage over all models evaluated on the full set of tasks.

\section{Appendix: Detailed Challenges with Developing Reliable and Minimal Long Reasoning Evaluations}
\label{app:related}

As mentioned in Section~\ref{sec:related}, there are a few pre-existing evaluations which study arbitrarily long reasoning tasks, with evaluations sometimes going up to 1M context, including \cite{zhang2024infty, hsieh2024rulerwhatsrealcontext, li2024needlebenchllmsretrievalreasoning,kuratov2024babilongtestinglimitsllms, loft}. These existing evaluations tend to suffer from at least one of a few deficiencies:
\begin{itemize}
    \item \textbf{Short-circuiting:} When an evaluation task does not account for information in either training data or later in the context which the model can use to avoid requiring the full context to answer a query, we say the model can ``short-circuit'' the long-context and perform well on the task while not actually representing a utilization of the context. As one example, the tasks in \cite{kuratov2024babilongtestinglimitsllms} may all suffer from this problem, both because the evaluation from which the expanded tasks are derived is heavily leaked (as it is a famous evaluation from almost a decade ago), and also because tasks themselves have built-in biases which remove the requirement that the model must use the full input \citep{kaushik-lipton-2018-much}. Some evaluations make efforts to combat this behavior. As an example, \cite{zhang2024infty} attempts to resolve this problem for a modified long QA task by swapping the names of characters in a book with randomly generated character name (``key entity replacement''). However, this mitigation does not control for the vast amounts of memorized information from pretraining which can still be helpful for localizing answers to questions about the book, independent of the character name - characters are often defined by many other quite distinguishable properties. The QA task from \cite{hsieh2024rulerwhatsrealcontext} has a similar problem in that it utilizes existing likely leaked questions from SQuAD and HotpotQA, augmented with filler context. Note that there are also similar issues with SQuAD \citep{kaushik-lipton-2018-much}. Finally, the authors of \cite{li2024needlebenchllmsretrievalreasoning} also note a flaw of this nature, as the needles are also derived from datasets curated from Wikipedia (and the datasets themselves are also present online, and thus likely leaked) - as they note, it is challenging to determine whether the models are simply utilizing internal knowledge here.
    \item \textbf{Secret retrieval tasks:} Some recently popular evaluation benchmarks (for instance, \cite{li2024needlebenchllmsretrievalreasoning, hsieh2024rulerwhatsrealcontext}) describe a subset of their tasks as corresponding to testing long-context reasoning capabilities. However, due to their construction, they often do not represent a task which requires anything more than single-needle or multi-needle retrieval. As one example of such a task, we consider the Variable Tracing (VT) task from RULER \citep{hsieh2024rulerwhatsrealcontext}. On the surface level, the task contains a context filled with many variable assignments (e.g. $X = 3; Y = X; Z = Y$ and so on, interspersed with distractor context). The model is then required to enumerate all variables which have the value $3$. However, in the default RULER implementation of this task, every variable which has been introduced in the context actually indeed has the value $3$ - there are no spurious variables present. Thus this setting of the task is ultimately reduced to a multi-needle retrieval task, where the needles correspond to the mentioned variable names, and is why we see exceedingly high performance on this task from many models. Related issues exist for the Common/Frequent Words Extraction (CWE/FWE) tasks in RULER \citep{hsieh2024rulerwhatsrealcontext}) -- if there are large disparities in the frequencies of the most common words, the model need not solve the task by actually performing fine-grained counting. Instead, it needs to simply generate the most likely continuous tokens (and repeats in the context increase the probability of a word being selected as the next token). Thus, the problem ultimately reduces to a task similar to multi-needle retrieval - which of the most common words have not yet been output? The GitHub repository for RULER acknowledges that while the tasks are configurable (and indeed, it is fairly easy to change the task parameters to be harder and require reasoning beyond retrieval), the benchmark itself avoids doing so since most models can attain a strong performance on the default task settings. The GitHub repository notes that more difficult variants of the tasks were not stress-tested, leading to the fairly high-scores on the RULER leaderboard. Ultimately, this trade-off results in the present form of RULER only testing single and multi-needle retrieval capabilities. We also note that the harder variants of VT/CWE/FWE all fit into the LSQ framework, but are significantly more artificial as compared to the \evalname tasks. Finally, many tasks which may require complex reasoning reduce to retrieval tasks if part of the prompt decomposes the problem into a sequence of retrieval problems, like the recently proposed HashHop task \citep{magic2024hashhop}. In HashHop, the model is required to output a key following a sequence of $k$ ``hops'' from key to value repeatedly in a large dictionary of keys and values which occupy the same vocabulary set. Assuming the model follows instructions or can follow few-shot prompting (possibly induced by training on the task), given examples where the model chains the hops together, the task reduces to computing the minimum over $k$ single-needle retrieval tasks (which is simply a more stringent metric than an average, but is not fundamentally different in what it measures compared to single-needle evaluations, and is furthermore not different in value as well if a model has a perfect single-needle retrieval performance). HashHop also fits into the LSQ framework and is more artificial than the \evalname tasks.
    \item \textbf{Out of distribution distractor context:} Many existing tasks proceed by inserting dramatically out-of-distribution context in order to form the irrelevant information component of the evaluation, such as Paul Graham essays or repeated phrases \citep{hsieh2024rulerwhatsrealcontext, li2024needlebenchllmsretrievalreasoning, needle}. This makes the problem significantly easier, since it implicitly brings any long reasoning task closer to a retrieval task -- if the relevant information is \textit{a priori} identifiable without understanding anything about the interrelated nature of the relevant information, then the task becomes effectively a multi-needle retrieval task.
    %\item \textbf{Explicit vs. implicit tasks:} Many tasks which have been previously proposed (for instance, \cite{nangia2018listops}) have the property that the context fully explicitly describes the information structure being queried by the model. Consider for example the task of reversing a presented list. This kind of task has two potential issues: 1) the task complexity increases as a function of the context size, in the sense that the object of interest that we wish the model to understand grows in complexity as a function of the context length; and 2) the task itself may be easier for the model to solve, as it may now be solvable via a specific form of iterated retrieval and copying. In many real-world settings, it is not necessarily the case that questions we may have about a topic we wish to understand are presented so explicitly in the text.
    \item \textbf{Training on toy synthetic tasks:} Some evaluation methods (often older evaluation sets from the pre-modern LLM era, like the Long Range Arena (LRA) \citep{tay2020long}) require the model to train on the task to see good performance on the task. Results on the non-CoT version of HashHop \citep{magic2024hashhop} were recently presented using this paradigm as well. We emphasize that the goal of these evaluation sets is rather different from the goal of Michelangelo: we would like to test for the ability of the model to learn useful reasoning circuits which operate over long-context which are very general. When these train-test evaluation sets impose a requirement to train on a highly idiosyncratic task, we are instead testing the ability of a given architecture to \textit{learn} a specific task, which while potentially interesting, is not the key point of interest in our setting, as we hope that via next-token prediction we will learn far more interesting long reasoning circuits that apply more generally over language and multimodal elements.
\end{itemize}

Synthetic evaluations do not have the above issues when constructed carefully, simply because their artificial nature allows for significantly more control in design. However, artificiality is itself undesirable. We would like to strike a middle ground between controllable (no short-circuiting, arbitrary context length, arbitrary controlled complexity) highly synthetic and unrealistic evaluations, and extremely realistic evaluations which are very difficult to control for. In particular, many issues arise: if a new long evaluation attempts to modify an existing leaked evaluation to make it longer context, then it suffers from the fact that the base evaluation is leaked. Often, inserted context has leaked information, leading to uncontrolled information presence and possible short-circuiting in-context. In completely realistic context this is also an issue since it gets proportionally more difficult as a function of context length to verify which information is present as well as where the information is present in the context. Realistic context evaluations are also not automatically extendable and require significant human labor, which can be error-prone due to the difficulty of parsing information in long-contexts for humans. This problem is exacerbated by the fact that for realistic tasks it is significantly more difficult to control the amount of relevant information and thus the complexity of the task. Ultimately, we would like a long-context evaluation to have both properties - it should be highly controllable (in terms of task complexity and context length), guarantee that the contexts for each task instance have no presence in pretraining data, and genuinely require the model to reason across the whole context, while not being an extremely toy or synthetic task.

Evaluations which are overly synthetic can express this nature in a few different ways: 1) the distractor context is highly synthetic and different from the relevant information (and is sometimes easily found available online); 2) the language of the task is itself highly unnatural (for instance - toy reasoning tasks based on alphabetic strings); 3) the task itself is highly contrived and it is unclear how it might be relevant. The last issue is particularly hard to avoid.

\end{document}